\documentclass[10pt,journal,compsoc]{IEEEtran}

\usepackage{graphicx}
\usepackage{amsmath}
\usepackage{amssymb}
\usepackage{booktabs}

\usepackage{multirow}
\usepackage{xcolor}
\usepackage{array}
\usepackage{multirow}
\usepackage{mathtools}
\usepackage{overpic}
\usepackage{stmaryrd}
\usepackage{url}
\usepackage{xspace}
\usepackage[ruled]{algorithm2e}

\usepackage{hyperref}

\usepackage[capitalize]{cleveref}
\crefname{section}{Sec.}{Secs.}
\Crefname{section}{Section}{Sections}
\crefname{table}{Tab.}{Tabs.}
\Crefname{table}{Table}{Tables}
\crefname{figure}{Fig.}{Figs.}
\Crefname{figure}{Figure}{Figures}
\crefname{equation}{Eq.}{Eqs.}
\Crefname{equation}{Equation}{Equations}
\crefname{appendix}{Appx.}{Appxs.}
\Crefname{Appendix}{Appendix}{Appendices}
\crefname{algorithm}{Alg.}{Algs.}
\Crefname{algorithm}{Algorithm}{Algorithms}

\ifCLASSOPTIONcompsoc
  \usepackage[nocompress]{cite}
\else
  \usepackage{cite}
\fi

\DeclareMathOperator*{\argmin}{\arg\!\min}

\makeatletter
\DeclareRobustCommand\onedot{\futurelet\@let@token\@onedot}
\def\@onedot{\ifx\@let@token.\else.\null\fi\xspace}

\def\eg{\emph{e.g}\onedot}

\def\ie{\emph{i.e}\onedot}

\def\vs{\emph{vs}\onedot}

\makeatother

\SetKwInOut{KwParam}{Parameter}

\newcommand{\tight}[1]{\hspace{1pt}{#1}{\hspace{1pt}}}
\newcommand{\medium}[1]{\hspace{2pt}{#1}{\hspace{2pt}}}

\newcolumntype{P}[1]{>{\centering\arraybackslash}p{#1}}
\newlength{\wdth}

\newcommand{\ptitle}[1]{\noindent\textbf{#1}\hspace{5pt}}

\newcommand{\paraspace}{\vspace{0pt}}

\begin{document}

\title{GeoTransformer: Fast and Robust Point Cloud Registration with Geometric Transformer}
%
%
%
%

\author{
Zheng~Qin, Hao~Yu, Changjian~Wang, Yulan~Guo, Yuxing~Peng, Slobodan~Ilic, Dewen~Hu, Kai~Xu$^{*}$
\IEEEcompsocitemizethanks{
\IEEEcompsocthanksitem Z.~Qin, C.~Wang, Y.~Guo, Y.~Peng, D.~Hu and K.~Xu are with National University of Defense Technology, China. Y.~Guo is also with Sun Yat-sen University, China. H.~Yu and S.~Ilic are with Technical University of Munich, Germany. S.~Ilic is also with Siemens AG, Germany.
\IEEEcompsocthanksitem Corresponding authors: Dewen~Hu (dwhu@nudt.edu.cn), Kai~Xu (kevin.kai.xu@gmail.com).
}
\thanks{Manuscript received April 19, 2005; revised August 26, 2015.}}

%
%

\markboth{Journal of \LaTeX\ Class Files,~Vol.~14, No.~8, August~2015}%
{Qin \MakeLowercase{\textit{et al.}}: Learning Fast and Robust Point Cloud Registration with Geometric Transformer}
%



\IEEEtitleabstractindextext{%

\begin{abstract}
We study the problem of extracting accurate correspondences for point cloud registration.
Recent keypoint-free methods have shown great potential through bypassing the detection of repeatable keypoints which is difficult to do especially in low-overlap scenarios. They seek correspondences over downsampled superpoints, which are then propagated to dense points.
Superpoints are matched based on whether their neighboring patches overlap.
Such sparse and loose matching requires contextual features capturing the geometric structure of the point clouds. We propose Geometric Transformer, or GeoTransformer for short, to learn geometric feature for robust superpoint matching. It encodes pair-wise distances and triplet-wise angles, making it invariant to rigid transformation and robust in low-overlap cases.
The simplistic design attains surprisingly high matching accuracy such that no RANSAC is required in the estimation of alignment transformation, leading to $100$ times acceleration.
Extensive experiments on rich benchmarks encompassing indoor, outdoor, synthetic, multiway and non-rigid demonstrate the efficacy of GeoTransformer.
Notably, our method improves the inlier ratio by $18{\sim}31$ percentage points and the registration recall by over $7$ points on the challenging 3DLoMatch benchmark.
Our code and models are available at \url{https://github.com/qinzheng93/GeoTransformer}.
\end{abstract}


}

\maketitle

\IEEEdisplaynontitleabstractindextext

%
\IEEEpeerreviewmaketitle


\IEEEraisesectionheading{\section{Introduction}\label{sec:intro}}
\IEEEPARstart{P}{oint} cloud registration is a fundamental task in graphics, vision and robotics. Given two partially overlapping 3D point clouds, the goal is to estimate a rigid transformation that aligns them. The problem has gained renewed interest recently thanks to the fast growing of 3D point representation learning and differentiable optimization.

The recent advances have been dominated by learning-based, correspondence-based methods~\cite{deng2018ppfnet,gojcic2019perfect,choy2019fully,bai2020d3feat,huang2021predator,yu2021cofinet}. A neural network is trained to extract point correspondences between two input point clouds, based on which an alignment transformation is calculated with a robust estimator, \eg, RANSAC. Most correspondence-based methods rely on keypoint detection~\cite{choy2019fully,bai2020d3feat,ao2021spinnet,huang2021predator}. However, it is challenging to detect repeatable keypoints across two point clouds, especially when they have small overlapping area. This usually results in low inlier ratio in the putative correspondences.

Inspired by the recent advances in image matching~\cite{rocco2018neighbourhood,zhou2021patch2pix,sun2021loftr}, keypoint-free methods~\cite{yu2021cofinet} downsample the input point clouds into superpoints and then match them through examining whether their local neighborhood (patch) overlaps. Such superpoint (patch) matching is then propagated to individual points, yielding dense point correspondences. Consequently, the accuracy of dense point correspondences highly depends on that of superpoint matches.


\begin{figure}[t]
  \begin{overpic}[width=1.0\linewidth]{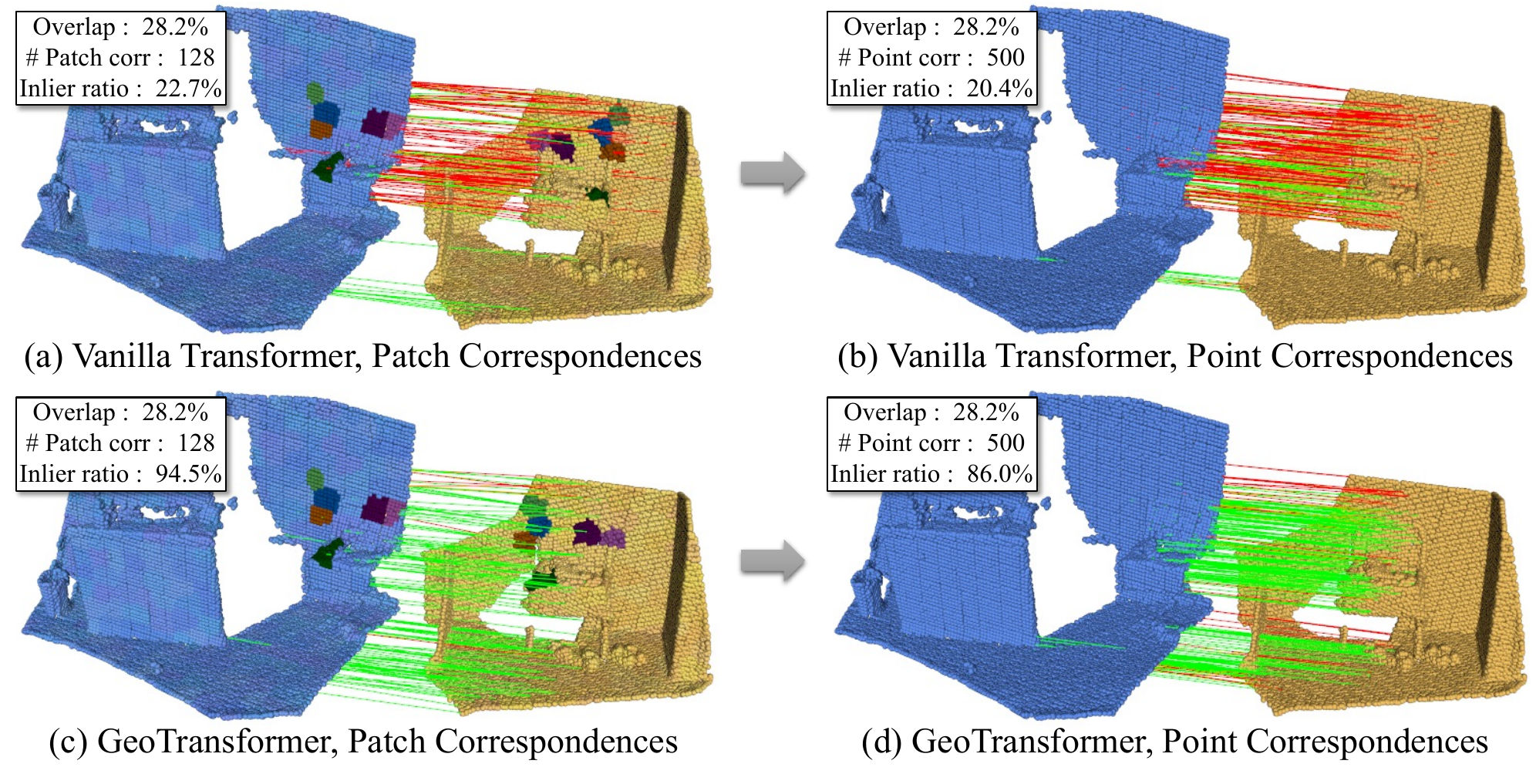}
  \end{overpic}
  \caption{Given two low-overlap point clouds, GeoTransformer improves inlier ratio over vanilla transformer significantly, both for superpoint (patch) level (left) and for dense point level (right). A few representative patch correspondences are visualized with distinct colors. Notice how GeoTransformer preserves the spatial consistency of the matching patches across two point clouds. It corrects the wrongly matched patches around the symmetric corners of the chair back (see the yellow point cloud).}
  \label{fig:teaser}
\end{figure}

Superpoint matching is sparse and loose. The upside is that it reduces strict point matching into loose patch overlapping, thus relaxing the repeatability requirement. Meanwhile, patch overlapping is a more reliable and informative constraint than distance-based point matching for learning correspondence; consider that two spatially close points could be geodesically distant. On the other hand, superpoint matching calls for features capturing more global context.

To this end, Transformer~\cite{vaswani2017attention} has been adopted~\cite{wang2019deep,yu2021cofinet} to encode contextual information in point cloud registration. However, vanilla transformer overlooks the geometric structure of the point clouds, which makes the learned features geometrically less discriminative and induces numerous outlier matches (\cref{fig:teaser}(top)). Although one can inject positional embeddings~\cite{zhao2021point,yang2019modeling}, the coordinate-based encoding is transformation-variant, which is problematic when registering point clouds given in arbitrary poses. We advocate that a point transformer for registration task should be learned with the \emph{geometric structure} of the point clouds so as to extract transformation-invariant geometric features. We propose \emph{Geometric Transformer}, or \emph{GeoTransformer} for short, for 3D point clouds which encodes only distances of point pairs and angles in point triplets.

Given a superpoint, we learn a non-local representation through geometrically ``pinpointing'' it w.r.t. all other superpoints based on pair-wise distances and triplet-wise angles. Self-attention mechanism is utilized to weigh the importance of those anchoring superpoints. Since distances and angles are invariant to rigid transformation, GeoTransformer learns geometric structure of point clouds efficiently, leading to highly robust superpoint matching even in low-overlap scenarios. \cref{fig:teaser}(left) demonstrates that GeoTransformer significantly improves the inlier ratio of superpoint (patch) correspondences. For better convergence, we devise an overlap-aware circle loss to make GeoTransformer focus on superpoint pairs with higher patch overlap.

Benefitting from the high-quality superpoint matches, our method attains high-inlier-ratio dense point correspondences (\cref{fig:teaser}(right)) using an optimal transport layer~\cite{sarlin2020superglue}, as well as highly robust and accurate registration without relying on RANSAC. Therefore, the registration part of our method runs extremely fast, \eg, $0.01$s for two point clouds with $5$K correspondences, $100$ times faster than RANSAC. Extensive experiments on indoor, outdoor, synthetic, multiway and non-rigid benchmarks~\cite{zeng20173dmatch,huang2021predator,geiger2012we,choi2015robust,li2022lepard} have demonstrated the efficacy of GeoTransformer. Specifically, our method attains significant improvements on challenging scenarios with low overlap and large rotations. For example, our method improves the inlier ratio by $18{\sim}31$ percentage points and the registration recall by over $7$ points on the 3DLoMatch benchmark~\cite{huang2021predator}. Our main contributions are:
\begin{itemize}
  \item A fast and accurate point cloud registration method which is both keypoint-free and RANSAC-free.
  \item A geometric transformer architecture which learns transformation-invariant geometric representation of point clouds for robust superpoint matching.
  \item An overlap-aware circle loss which reweights the loss of each superpoint match according to the patch overlap ratio for better convergence.
\end{itemize}

A previous version of this work was published at CVPR 2022~\cite{qin2022geometric}. This paper extends the conference version with the following new contributions: First, to reduce the memory footprint and the computational cost of GeoTransformer, we propose \emph{shared geometric self-attention} which makes the attention weights for the geometric structure embeddings shared across all self-attention modules. Second, we extend our method to deal with non-rigid registration through relaxing the selection of superpoint correspondences, demonstrating the strong generality of GeoTransformer. Third, we further conduct more extensive experiments and detailed ablation analysis to provide a thorough understanding of the effectiveness of GeoTransformer.


\section{Related Work}
\label{sec:related}

\ptitle{Correspondence-based methods.}
Our work follows the line of the correspondence-based methods~\cite{deng2018ppfnet,deng2018ppf,gojcic2019perfect,choy2019fully}. They first extract correspondences between two point clouds and then recover the transformation with robust pose estimators, \eg, RANSAC. Thanks to the robust estimators, they achieve state-of-the-art performance in indoor and outdoor scene registration. These methods can be further categorized into two classes according to how they extract correspondences. The first class aims to detect more repeatable keypoints~\cite{bai2020d3feat,huang2021predator} and learn more powerful descriptors for the keypoints~\cite{choy2019fully,ao2021spinnet,wang2022you}. While the second class~\cite{yu2021cofinet} retrieves correspondences without keypoint detection by considering all possible matches. Our method follows the detection-free methods and improves the accuracy of correspondences by leveraging the geometric information.

\paraspace
\ptitle{Direct registration methods.}
Recently, direct registration methods have emerged. They estimate the transformation with a neural network in an end-to-end manner and eliminate the use of a robust estimator. According to how the alignment transformation is estimated, these methods can be further classified into two classes. The first class~\cite{wang2019deep,wang2019prnet,yuan2020deepgmr,li2020iterative,yew2020rpm,fu2021robust,zhang2022end} follows the idea of ICP~\cite{besl1992method}, which iteratively establishes soft correspondences and computes the transformation with differentiable weighted SVD. And the second class~\cite{aoki2019pointnetlk,huang2020feature,xu2021omnet,xu2022finet} first extracts a global feature vector for each point cloud and regresses the transformation with the global feature vectors. Due to the lack of a robust estimator, direct registration methods opt to adopt an iterative registration scheme to progressively refine the estimated transformation. Albeit achieving promising results on single synthetic shapes, direct registration methods could still fail in large-scale scenes as stated in~\cite{huang2021predator}.

\paraspace
\ptitle{Deep robust estimators.}
As traiditional robust estimators such as RANSAC suffer from slow convergence and instability in case of high outlier ratio, deep robust estimators~\cite{pais20203dregnet,choy2020deep,bai2021pointdsc} have been proposed as the alternatives for them. They usually contain a classification network to reject outliers and an estimation network to compute the transformation. Compared with traditional robust estimators, they achieve improvements in both accuracy and speed. However, they require training a specific network. In comparison, our method achieves fast and accurate registration with a parameter-free local-to-global registration scheme.

\paraspace
\ptitle{Geometric consistency in point cloud registration.}
Geometric consistency has been an important and long-standing research topic in point cloud registration. Given two point clouds in arbitrary poses, certain geometric properties such as distances and angles are preserved between them, which provides a strong geometric guidance for registration. To this end, previous hand-crafted methods~\cite{rusu2009fast,drost2010model,raposo2017using} directly encode lengths and angles around an anchor point to obtain transformation-invariant descriptors. However, these descriptors are not aware of the global structure, which restricts their distinctiveness. Besides, geometric consistency has also been adopted to reject outlier correspondences such that more accurate transformation could be recovered~\cite{leordeanu2005spectral,yang2020teaser,bai2021pointdsc}. These methods need a preceding correspondence extractor and are orthogonal to this work.


\section{Method}
\label{sec:geotrans}


\begin{figure*}[t]
  \begin{overpic}[width=1.0\linewidth]{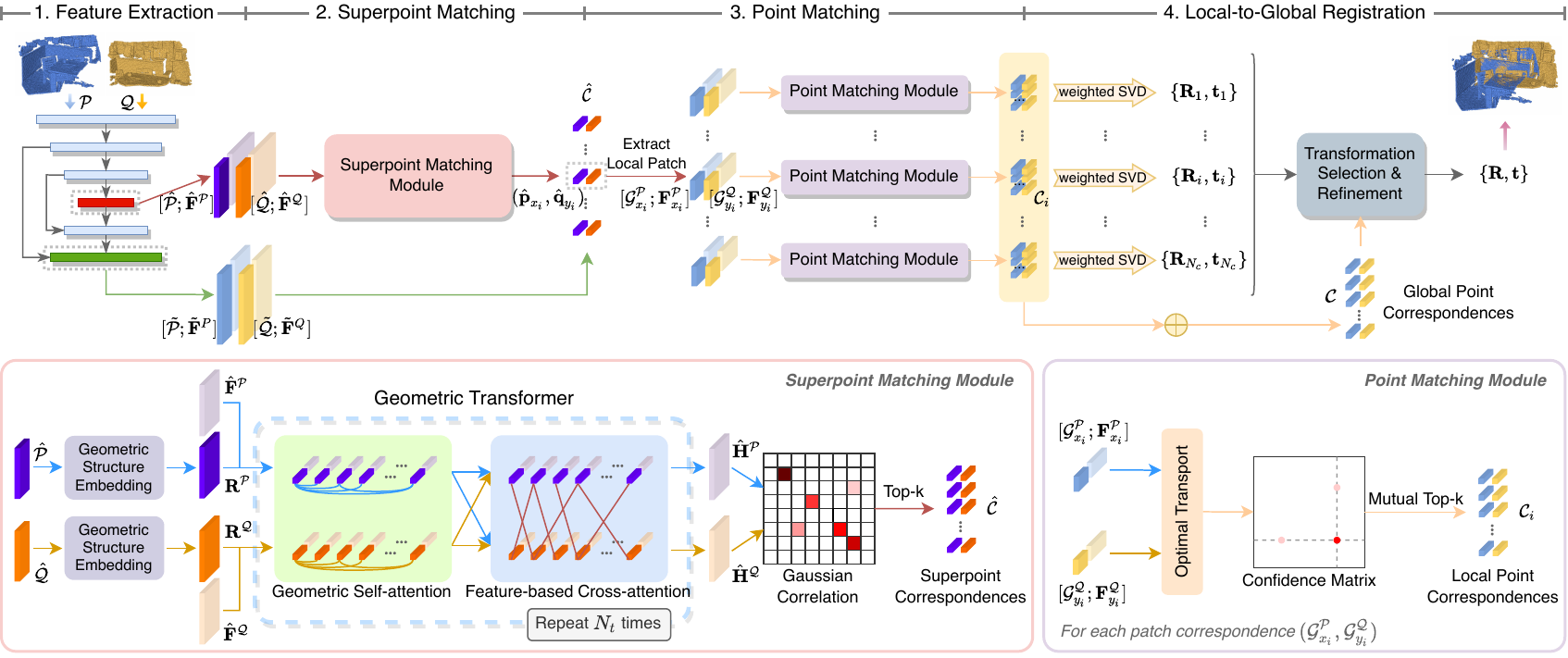}
  \end{overpic}
  \caption{\textbf{The overall pipeline of our method}. The backbone downsamples the input point clouds and learns features in multiple resolution levels. The Superpoint Matching Module extracts high-quality superpoint correspondences between $\hat{\mathcal{P}}$ and $\hat{\mathcal{Q}}$ using the Geometric Transformer which iteratively encodes intra-point-cloud geometric structures and inter-point-cloud geometric consistency. The superpoint correspondences are then propagated to dense points $\tilde{\mathcal{P}}$ and $\tilde{\mathcal{Q}}$ by the Point Matching Module. Finally, the transformation is computed with a local-to-global registration method.}
  \label{fig:overview}
\end{figure*}

Given two point clouds $\mathcal{P} = \{\textbf{p}_i \in \mathbb{R}^3 \mid i = 1, ..., N\}$ and $\mathcal{Q} = \{\textbf{q}_i \in \mathbb{R}^3 \mid i = 1, ..., M\}$, our goal is to estimate a rigid transformation $\textbf{T} = \{\textbf{R}, \textbf{t}\}$ which aligns the two point clouds, with a 3D rotation $\textbf{R} \in \mathcal{SO}(3)$ and a 3D translation $\textbf{t} \in \mathbb{R}^3$. The transformation can be solved by:
\begin{equation}
\min_{\textbf{R}, \textbf{t}} \sum\nolimits_{(\textbf{p}^{*}_{x_i}, \textbf{q}^{*}_{y_i}) \in \mathcal{C}^{*}} \lVert \textbf{R} \cdot \textbf{p}^{*}_{x_i} + \textbf{t} - \textbf{q}^{*}_{y_i} \rVert^2_2.
\end{equation}
Here $\mathcal{C}^{*}$ is the set of ground-truth correspondences between $\mathcal{P}$ and $\mathcal{Q}$. Since $\mathcal{C}^{*}$ is unknown in reality, we need to first establish point correspondences between two point clouds and then estimate the alignment transformation.

Our method adopts the hierarchical correspondence paradigm which finds correspondences in a coarse-to-fine manner. We adopt KPConv-FPN to simultaneously downsample the input point clouds and extract point-wise features (\cref{sec:model-backbone}). The first and the last (coarsest) level downsampled points correspond to the dense points and the superpoints to be matched. A \emph{Superpoint Matching Module} is used to extract superpoint correspondences whose neighboring local patches overlap with each other (\cref{sec:model-pam}). Based on that, a \emph{Point Matching Module} then refines the superpoint correspondences to dense points (\cref{sec:model-pom}). At last, the alignment transformation is recovered from the dense correspondences without relying on RANSAC (\cref{sec:model-estimation}). The pipeline is illustrated in \cref{fig:overview}.

\subsection{Superpoint Sampling and Feature Extraction}
\label{sec:model-backbone}

We utilize the KPConv-FPN backbone~\cite{thomas2019kpconv,lin2017feature} to extract multi-level features for the point clouds. A byproduct of the point feature learning is point downsampling. We work on downsampled points since point cloud registration can actually be pinned down by the correspondences of a much coarser subset of points. The original point clouds are usually too dense so that point-wise correspondences are redundant
and sometimes too clustered to be useful.

The points correspond to the coarsest resolution, denoted by $\hat{\mathcal{P}}$ and $\hat{\mathcal{Q}}$, are treated as \emph{superpoints} to be matched. The associated learned features are denoted as $\hat{\textbf{F}}{}^{\mathcal{P}} \medium{\in} \mathbb{R}^{\lvert \hat{\mathcal{P}} \rvert \times \hat{d}}$ and $\hat{\textbf{F}}{}^{\mathcal{Q}} \medium{\in} \mathbb{R}^{\lvert \hat{\mathcal{Q}} \rvert \times \hat{d}}$. The dense point correspondences are computed at $1/2$ of the original resolution, \ie, the first level downsampled points denoted by $\tilde{\mathcal{P}}$ and $\tilde{\mathcal{Q}}$. Their learned features are represented by $\tilde{\textbf{F}}{}^{\mathcal{P}} \medium{\in} \mathbb{R}^{\lvert \tilde{\mathcal{P}} \rvert \times \tilde{d}}$ and $\tilde{\textbf{F}}{}^{\mathcal{Q}} \medium{\in} \mathbb{R}^{\lvert \tilde{\mathcal{Q}} \rvert \times \tilde{d}}$.


\begin{figure}[t]
  \centering
  \begin{overpic}[width=1.0\linewidth]{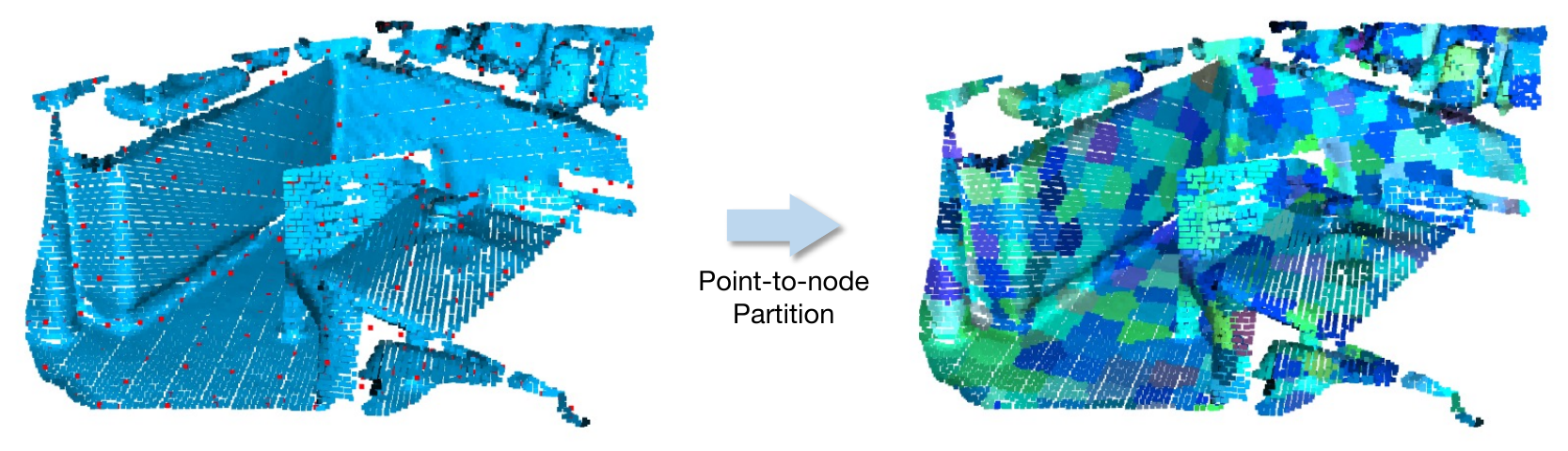}
  \end{overpic}
  \caption{
  \textbf{Point-to-node grouping strategy}. Each point is assigned to its nearest superpoint.
  Left: the point cloud (in blue) and the sampled superpoints (in red). Right: the points are color-coded according to the superpoints that they are assigned to.
  }
  \label{fig:partition}
\end{figure}

For each superpoint, we construct a local \emph{patch} of points around it using the point-to-node grouping strategy \cite{li2018so,yu2021cofinet}. As shown in \cref{fig:partition}, each point in $\tilde{\mathcal{P}}$ and its features from $\tilde{\textbf{F}}{}^{\mathcal{P}}$ are assigned to its nearest superpoint in the geometric space:
\begin{equation}
\mathcal{G}^{\mathcal{P}}_i = \{\tilde{\textbf{p}} \in \tilde{\mathcal{P}} \mid i = \argmin\nolimits_j(\lVert \tilde{\textbf{p}} - \hat{\textbf{p}}_j \rVert_2), \hat{\textbf{p}}_j \in \hat{\mathcal{P}}\}.
\end{equation}
This essentially leads to a Voronoi decomposition of the input point cloud seeded by superpoints. The feature matrix associated with the points in $\mathcal{G}^{\mathcal{P}}_i$ is denoted as $\textbf{F}^{\mathcal{P}}_i \subset \tilde{\textbf{F}}{}^{\mathcal{P}}$. The superpoints with an empty patch are removed. The patches $\{\mathcal{G}^{\mathcal{Q}}_i\}$ and the feature matrices $\{\textbf{F}^{\mathcal{Q}}_i\}$ for $\mathcal{Q}$ are computed and denoted in a similar way.

\subsection{Superpoint Matching Module}
\label{sec:model-pam}


\begin{figure}[t]
  \centering
  \begin{overpic}[width=1.0\linewidth]{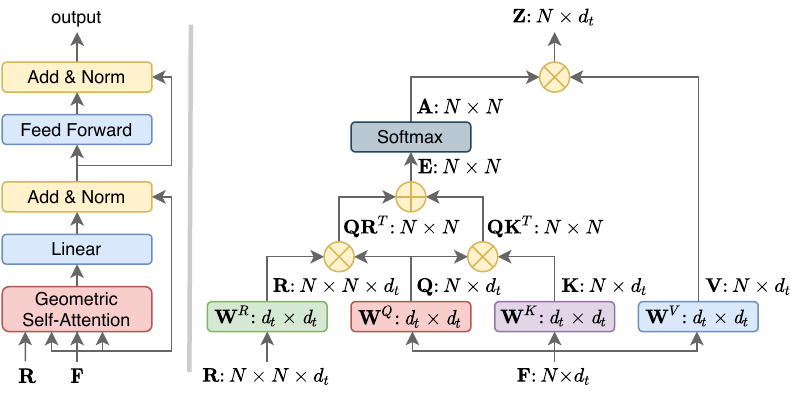}
  \end{overpic}
  \caption{\textbf{Geometric self-attention module}.
  Left: The structure of geometric self-attention module. Right: The computation graph of geometric self-attention mechanism.
  }
  \label{fig:geotr}
\end{figure}

\ptitle{Geometric Transformer.}
Global context has proven critical in many computer vision tasks~\cite{dosovitskiy2020image,sun2021loftr,yu2021cofinet}. For this reason, transformer has been adopted to leverage global contextual information for point cloud registration. However, existing methods~\cite{wang2019deep,huang2021predator,yu2021cofinet} usually feed transformer with only high-level point cloud features and does not explicitly encode the geometric structure. This makes the learned features geometrically less discriminative, which causes severe matching ambiguity and numerous outlier matches, especially in low-overlap cases. A straightforward recipe is to explicitly inject positional embeddings~\cite{yang2019modeling,zhao2021point} of 3D point coordinates. However, the resultant coordinate-based transformers are naturally \emph{transformation-variant}, while registration requires \emph{transformation invariance} since the input point clouds can be in arbitrary poses.

To this end, we propose \emph{Geometric Transformer} which not only encodes high-level point features but also explicitly captures intra-point-cloud geometric structures and inter-point-cloud geometric consistency. GeoTransformer is composed of a \emph{geometric self-attention} module for learning intra-point-cloud features and a \emph{feature-based cross-attention} module for modeling inter-point-cloud consistency. The two modules are interleaved for $N_t$ times to extract hybrid features $\hat{\textbf{H}}{}^{\mathcal{P}}$ and $\hat{\textbf{H}}{}^{\mathcal{Q}}$ for reliable superpoint matching (see \cref{fig:overview} (bottom left)).

\paraspace
\ptitle{Geometric self-attention.}
We design a \emph{geometric self-attention} to learn the global correlations in both feature and geometric spaces among the superpoints for each point cloud. In the following, we describe the computation for $\hat{\mathcal{P}}$ and the same goes for $\hat{\mathcal{Q}}$. Given the input feature matrix $\textbf{X} \medium{\in} \mathbb{R}^{\lvert \hat{\mathcal{P}} \vert \times d_t}$, the output feature matrix $\textbf{Z} \medium{\in} \mathbb{R}^{\lvert \hat{\mathcal{P}} \vert \times d_t}$ is the weighted sum of all projected input features:
\begin{equation}
\textbf{z}_i = \sum_{j=1}^{\lvert \hat{\mathcal{P}} \vert} a_{i, j} (\textbf{x}_j\textbf{W}^V),
\end{equation}
where the weight coefficient $a_{i, j}$ is computed by a row-wise softmax on the attention score $e_{i, j}$, and $e_{i, j}$ is computed as:
\begin{equation}
e_{i, j} = \frac{(\textbf{x}_i\textbf{W}^Q)(\textbf{x}_j\textbf{W}^K + \textbf{r}_{i, j}\textbf{W}^R)^T}{\sqrt{d_{t}}}.
\label{eq:gsa}
\end{equation}
Here, $\textbf{r}_{i, j} \medium{\in} \mathbb{R}^{d_t}$ is a \emph{geometric structure embedding} to be described in the next. $\textbf{W}^Q, \textbf{W}^K, \textbf{W}^V, \textbf{W}^R \in \mathbb{R}^{d_t \times d_t}$ are the respective projection matrices for queries, keys, values and geometric structure embeddings. \cref{fig:geotr} shows the structure and the computation of geometric self-attention.


\begin{figure}[t]
  \begin{overpic}[width=1.0\linewidth]{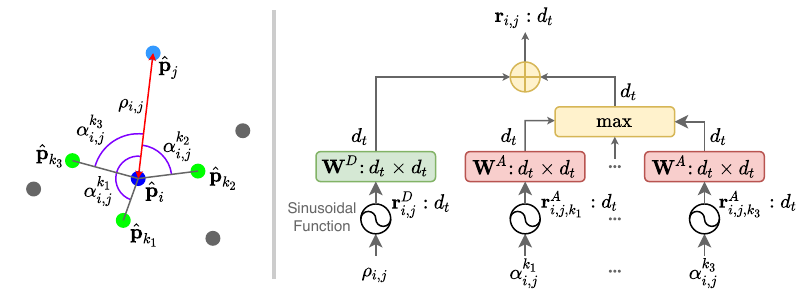}
  \end{overpic}
  \caption{
  \textbf{Geometric structure embedding}.
  Left: An illustration of the pair-wise distance and the triplet-wise angles encoded.
  Right: The computation graph of the geometric structure embedding.}
  \label{fig:rge}
\end{figure}

We design a novel \emph{geometric structure embedding} to encode the transformation-invariant geometric structure of the superpoints. The core idea is to leverage the distances and angles computed with the superpoints which are consistent across different point clouds of the same scene. Given two superpoints $\hat{\textbf{p}}_i, \hat{\textbf{p}}_j \tight{\in} \hat{\mathcal{P}}$, their geometric structure embedding consists of a \emph{pair-wise distance embedding} and a \emph{triplet-wise angular embedding}, which will be described below.

(1) \emph{Pair-wise Distance Embedding}.
Given the distance $\rho_{i, j} \hspace{1pt} {=} \hspace{1pt} \lVert \hat{\textbf{p}}_i - \hat{\textbf{p}}_j \rVert_2$ between $\hat{\textbf{p}}_i$ and $\hat{\textbf{p}}_j$, the distance embedding $\textbf{r}^D_{i, j} \in \mathbb{R}^{d_t}$ between them is computed by applying a sinusoidal function \cite{vaswani2017attention} on $\rho_{i, j}$:
\begin{equation}
\left\{
\begin{aligned}
r^D_{i, j, 2k} & = \sin(\frac{d_{i, j} / \sigma_d}{10000^{2k / d_t}}) \\
r^D_{i, j, 2k+1} & = \cos(\frac{d_{i, j} / \sigma_d}{10000^{2k / d_t}})
\end{aligned},
\right.
\label{eq:pde}
\end{equation}
where $\sigma_d$ is a temperature hyper-parameter used to tune the sensitivity on distance variations.

(2) \emph{Triplet-wise Angular Embedding}.
We compute angular embedding with triplets of superpoints. We first select the $k$ nearest neighbors $\mathcal{K}_i$ of $\hat{\textbf{p}}_i$.
For each $\hat{\textbf{p}}_x \tight{\in} \mathcal{K}_i$, we compute the angle $\alpha^x_{i,j} \tight{=} \angle(\Delta_{x, i}, \Delta_{j, i})$, where $\Delta_{i, j} \tight{:=} \hat{\textbf{p}}_i \hspace{1pt} {-} \hspace{1pt} \hat{\textbf{p}}_j$. The triplet-wise angular embedding $\textbf{r}^A_{i, j, x} \in \mathbb{R}^{d_t}$ is then computed with a sinusoidal function on $\alpha^x_{i,j}$:
\begin{equation}
\left\{
\begin{aligned}
r^A_{i, j, x, 2l} & = \sin(\frac{\alpha^x_{i, j} / \sigma_a}{10000^{2l / d_t}}) \\
r^A_{i, j, x, 2l+1} & = \cos(\frac{\alpha^x_{i, j} / \sigma_a}{10000^{2l / d_t}})
\end{aligned},
\right.
\label{eq:tae}
\end{equation}
where $\sigma_a$ controls the sensitivity on angular variations.

Finally, the geometric structure embedding $\textbf{r}_{i, j}$ is computed by aggregating the pair-wise distance embedding and the triplet-wise angular embedding:
\begin{equation}
\textbf{r}_{i, j} = \textbf{r}^D_{i, j}\textbf{W}^D + {\max}_x\left\{\textbf{r}^A_{i, j, x}\textbf{W}^A\right\},
\label{eq:gse}
\end{equation}
where $\textbf{W}^D, \textbf{W}^A \in \mathbb{R}^{d_t \times d_t}$ are the respective projection matrices for the two types of embeddings. We use max pooling here to improve the robustness to the varying nearest neighbors of a superpoint due to self-occlusion. \cref{fig:rge} illustrates the computation of geometric structure embedding.


\begin{figure}[t]
  \centering
  \begin{overpic}[width=1.0\linewidth]{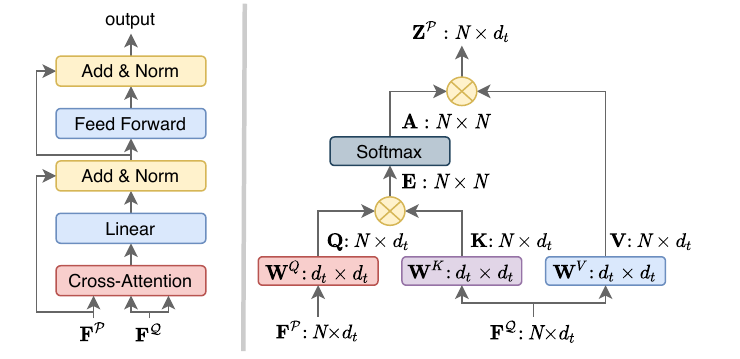}
  \end{overpic}
  \caption{
  \textbf{Feature-based cross-attention module}.
  Left: The structure of feature-based cross-attention module. Right: The computation graph of cross-attention mechanism.
  }
  \label{fig:cross-att}
\end{figure}

\paraspace
\ptitle{Shared geometric self-attention.}
Albeit enjoying a strong representation capability, the geometric self-attention suffers from the heavy computation of the embedding projection $\mathbf{r}_{i, j}\mathbf{W}^R$ in \cref{eq:gsa}. The computational complexity of the standard geometric self-attention is $O(\lvert \hat{\mathcal{P}} \rvert d_t^2 + \lvert \hat{\mathcal{P}} \rvert^2 d_t^2)$, which limits its scalability and efficiency especially when the number of superpoints is large. To reduce the computation, we design a \emph{shared geometric self-attention} which makes the projection weights $\mathbf{W}^R$ shared across all geometric self-attention modules and apply $\mathbf{W}^R$ in \cref{eq:gse} instead:
\begin{equation}
e_{i, j} = \frac{(\textbf{x}_i\textbf{W}^Q)(\textbf{x}_j\textbf{W}^K + \textbf{r}_{i, j})}{\sqrt{d_{t}}}.
\end{equation}
The geometric structure embedding is then computed as
\begin{equation}
\textbf{r}_{i, j} = \sigma(\textbf{r}^D_{i, j}\textbf{W}^D + {\max}_x\left\{\textbf{r}^A_{i, j, x}\textbf{W}^A\right\})\mathbf{W}^R,
\end{equation}
where $\sigma$ is the LeakyReLU function. With this modification, the computation complexity of geometric self-attention is reduced to $O(\lvert \hat{\mathcal{P}} \rvert d_t^2)$. As shown in \cref{sec:exp-indoor}, shared geometric self-attention attains comparable accuracy with the standard version with a significant reduction of computation time.

\paraspace
\ptitle{Feature-based cross-attention.}
Cross-attention is a typical module for point cloud registration task~\cite{huang2021predator,wang2019deep,yu2021cofinet}, used to perform feature exchange between two input point clouds. Given the self-attention feature matrices $\textbf{X}^{\mathcal{P}}$, $\textbf{X}^{\mathcal{Q}}$ for $\hat{\mathcal{P}}$, $\hat{\mathcal{Q}}$ respectively, the cross-attention feature matrix $\textbf{Z}^{\mathcal{P}}$ of $\hat{\mathcal{P}}$ is computed with the features of $\hat{\mathcal{Q}}$:
\begin{equation}
\textbf{z}^{\mathcal{P}}_i = \sum_{j=1}^{\lvert \hat{\mathcal{Q}} \rvert} a_{i, j} (\textbf{x}^{\mathcal{Q}}_j\textbf{W}^V).
\end{equation}
Similarly, $a_{i, j}$ is computed by a row-wise softmax on the cross-attention score $e_{i, j}$, and $e_{i, j}$ is computed as the feature correlation between the $\textbf{X}^{\mathcal{P}}$ and $\textbf{X}^{\mathcal{Q}}$:
\begin{equation}
e_{i, j} = \frac{(\textbf{x}^{\mathcal{P}}_i\textbf{W}^Q)(\textbf{x}^{\mathcal{Q}}_j\textbf{W}^K)^T}{\sqrt{d_{t}}}.
\end{equation}
\cref{fig:cross-att} shows the structure and the computation of the cross-attention. The cross-attention features for $\mathcal{Q}$ are computed in the same way. While the geometric self-attention module encodes the transformation-invariant geometric structure for each individual point cloud, the feature-based cross-attention module can model the geometric consistency across the two point clouds. The resultant hybrid features are both invariant to transformation and robust for reasoning correspondence.

\paraspace
\ptitle{Superpoint matching.}
To find the superpoint correspondences, we propose a matching scheme based on global feature correlation. We first normalize $\hat{\textbf{H}}{}^{\mathcal{P}}$ and $\hat{\textbf{H}}{}^{\mathcal{Q}}$ onto a unit hypersphere and compute a Gaussian correlation matrix $\textbf{S} \tight{\in} \mathbb{R}^{\lvert \hat{\mathcal{P}} \rvert \times \lvert \hat{\mathcal{Q}} \rvert}$ with $s_{i, j} \tight{=} \exp(-\lVert \hat{\textbf{h}}{}^{\mathcal{P}}_i \tight{-} \hat{\textbf{h}}{}^{\mathcal{Q}}_j\rVert_2^2)$. In practice, some patches of a point cloud are less geometrically discriminative and have numerous similar patches in the other point cloud. Besides our powerful hybrid features, we also perform a dual-normalization operation \cite{rocco2018neighbourhood,sun2021loftr} on $\textbf{S}$ to further suppress ambiguous matches, leading to $\bar{\textbf{S}}$ with
\begin{equation}
\bar{s}_{i, j} = \frac{s_{i, j}}{\sum_{k=1}^{\lvert \hat{\mathcal{Q}} \rvert} s_{i, k}} \cdot \frac{s_{i, j}}{\sum_{k=1}^{\lvert \hat{\mathcal{P}} \rvert} s_{k, j}}.
\end{equation}
We found that this suppression can effectively eliminate wrong matches. Finally, we select the largest $N_{c}$ entries in $\bar{\textbf{S}}$ as the \emph{superpoint correspondences}:
\begin{equation}
\hat{\mathcal{C}} = \{ (\hat{\textbf{p}}_{x_i}, \hat{\textbf{q}}_{y_i}) \mid (x_i, y_i) \in \mathrm{topk}_{x, y}(\bar{s}_{x, y}) \}.
\end{equation}
Due to the powerful geometric structure encoding of GeoTransformer, our method is able to achieve accurate registration in low-overlap cases and with less point correspondences, and most notably, in a RANSAC-free manner.

\subsection{Point Matching Module}
\label{sec:model-pom}

Having obtained the superpoint correspondences, we extract point correspondences using a simple yet effective \emph{Point Matching Module}. At point level, we use only local point features learned by the backbone. The rationale is that point level matching is mainly determined by the vicinities of the two points being matched, once the global ambiguity has been resolved by superpoint matching. This design choice improves the robustness.

For each superpoint correspondence $\hat{\mathcal{C}}_i = (\hat{\textbf{p}}_{x_i}, \hat{\textbf{q}}_{y_i})$, an optimal transport layer \cite{sarlin2020superglue} is used to extract the \emph{local} dense point correspondences between $\mathcal{G}^{\mathcal{P}}_{x_i}$ and $\mathcal{G}^{\mathcal{Q}}_{y_i}$. Specifically, we first compute a cost matrix $\textbf{C}_i \in \mathbb{R}^{n_i \times m_i}$:
\begin{equation}
\textbf{C}_i = \textbf{F}^{\mathcal{P}}_{x_i} (\textbf{F}^{\mathcal{Q}}_{y_i})^T / \sqrt{\tilde{d}},
\end{equation}
where $n_i = \lvert \mathcal{G}^{\mathcal{P}}_{x_i} \rvert$, $m_i = \lvert \mathcal{G}^{\mathcal{Q}}_{y_i} \rvert$. The cost matrix $\textbf{C}_i$ is then augmented into $\bar{\textbf{C}}_i$ by appending a new row and a new column as in \cite{sarlin2020superglue}, filled with a learnable dustbin parameter $\alpha$. We then utilize the Sinkhorn algorithm \cite{sinkhorn1967concerning} on $\bar{\textbf{C}}_i$ to compute a soft assignment matrix $\bar{\textbf{Z}}_i$ which is then recovered to $\textbf{Z}_i$ by dropping the last row and the last column. We use $\textbf{Z}_i$ as the confidence matrix of the candidate matches and extract point correspondences via mutual top-$k$ selection, where a point match is selected if it is among the $k$ largest entries of both the row and the column that it resides in:
\begin{equation}
\mathcal{C}_i \tight{=} \Bigl\{\bigl(\mathcal{G}^{\mathcal{P}}_{x_i}(x_j), \mathcal{G}^{\mathcal{Q}}_{y_i}(y_j)\bigr) \tight{\mid} (x_j, y_j) \tight{\in} \mathrm{mutual\_topk}_{x, y}(z^i_{x, y})\Bigr\}.
\end{equation}
The point correspondences computed from each superpoint match are then collected together to form the final \emph{global} dense point correspondences: $\mathcal{C} = \bigcup_{i=1}^{N_c} \mathcal{C}_i$.

\subsection{RANSAC-free Local-to-Global Registration}
\label{sec:model-estimation}


\begin{algorithm}[t]
\small
\caption{Local-to-Global Registration}
\label{alg:lgr}
\KwIn{$\mathcal{C}_i$: local point correspondences of superpoint correspondences}
\KwOut{$\mathbf{R}$, $\textbf{t}$: alignment transformation}
\emph{1. local step}\\
\For {$i \leftarrow 1, ..., N_c$} {
	Compute $\mathbf{R}_i$, $\mathbf{t}_i$ by solving \cref{eq:weighted-svd} on $\mathcal{C}_i$.
}
\emph{2. global step}\\
Select best transformation candidate $\mathbf{R}$, $\mathbf{t}$ by \cref{eq:candidate-selection}.\\ 
$\mathcal{C} \leftarrow \mathcal{C}_1 \cup ... \cup \mathcal{C}_{N_c}$\\
\For {$t \leftarrow 1, ..., N_r$} {
	${}^{(t)}\mathcal{C}$ $\leftarrow$ inliers in $\mathcal{C}$ under $\mathbf{R}$ and $\mathbf{t}$.\\
	Update $\mathbf{R}$, $\mathbf{t}$ by solving \cref{eq:weighted-svd} on ${}^{(t)}\mathcal{C}$.
}
\end{algorithm}

Previous methods generally rely on robust pose estimators to estimate the transformation since the putative correspondences are often predominated by outliers. Most robust estimators such as RANSAC suffer from slow convergence. Given the high inlier ratio of GeoTransformer, we are able to achieve robust registration without relying on robust estimators, which also greatly reduces computation cost.

We design a \emph{local-to-global registration} (LGR) scheme. As a hypothesize-and-verify approach, LGR is comprised of a local phase of transformation candidates generation and a global phase for transformation selection. In the local phase, we solve for a transformation $\textbf{T}_i \tight{=} \{\textbf{R}_i, \textbf{t}_i\}$ for each superpoint match using its \emph{local point correspondences}:
\begin{equation}
\textbf{R}_i, \textbf{t}_i = \min_{\textbf{R}, \textbf{t}} \sum\nolimits_{(\tilde{\textbf{p}}_{x_j}, \tilde{\textbf{q}}_{y_j}) \in \mathcal{C}_i} w^i_j \lVert \textbf{R} \cdot \tilde{\textbf{p}}_{x_j} + \textbf{t} - \tilde{\textbf{q}}_{y_j} \rVert_2^2.
\label{eq:weighted-svd}
\end{equation}
This can be solved in closed form using weighted SVD~\cite{besl1992method}. The corresponding confidence score for each correspondence in $\textbf{Z}_i$ is used as the weight $w^i_j$. Benefitting from the high-quality correspondences, the transformations obtained in this phase are already very accurate. In the global phase, we select the transformation which admits the most inlier matches over the entire \emph{global point correspondences}:
\begin{equation}
\textbf{R}, \textbf{t} = \max_{\textbf{R}_i, \textbf{t}_i} \sum\nolimits_{(\tilde{\textbf{p}}_{x_j}, \tilde{\textbf{q}}_{y_j}) \in \mathcal{C}} \llbracket \lVert \textbf{R}_i \cdot \tilde{\textbf{p}}_{x_j} + \textbf{t}_i - \tilde{\textbf{q}}_{y_j} \rVert_2 < \tau_a \rrbracket,
\label{eq:candidate-selection}
\end{equation}
where $\llbracket \cdot \rrbracket$ is the Iverson bracket. $\tau_a$ is the acceptance radius. We then iteratively re-estimate the transformation with the surviving inlier matches for $N_r$ times by solving \cref{eq:weighted-svd}.
\cref{alg:lgr} shows the computation of the local-to-global registration. As shown in \cref{sec:exp-indoor}, our approach achieves comparable registration accuracy with RANSAC but reduces the computation time by more than $100$ times. Moreover, unlike deep robust estimators~\cite{choy2020deep,pais20203dregnet,bai2021pointdsc}, our method is parameter-free and no network training is needed.

\subsection{Loss Functions}
\label{sec:model-loss}

The loss function $\mathcal{L} = \mathcal{L}_{oc} + \mathcal{L}_{p}$ is composed of an \emph{overlap-aware circle loss} $\mathcal{L}_{oc}$ for superpoint matching and a \emph{point matching loss} $\mathcal{L}_{p}$ for point matching.

\paraspace
\ptitle{Overlap-aware circle loss.}
Existing methods~\cite{sun2021loftr,yu2021cofinet} usually formulate superpoint matching as a multi-label classification problem and adopt a cross-entropy loss with dual-softmax~\cite{sun2021loftr} or optimal transport~\cite{sarlin2020superglue,yu2021cofinet}. Each superpoint is assigned (classified) to one or many of the other superpoints, where the ground truth is computed based on patch overlap and it is very likely that one patch could overlap with multiple patches. By analyzing the gradients from the cross-entropy loss, we find that the positive classes with high confidence scores are suppressed by positive gradients in the multi-label classification. This hinders the model from extracting reliable superpoint correspondences.

To address this issue, we opt to extract superpoint descriptors in a metric learning fashion. A straightforward solution is to adopt a circle loss~\cite{sun2020circle} similar to~\cite{bai2020d3feat,huang2021predator}. However, the circle loss overlooks the differences between the positive samples and weights them equally. As a result, it struggles in matching patches with relatively low overlap. For this reason, we design an \emph{overlap-aware circle loss} to focus the model on those matches with high overlap. We select the patches in $\mathcal{P}$ which have at least one positive patch in $\mathcal{Q}$ to form a set of anchor patches, $\mathcal{A}$. A pair of patches are positive if they share at least $10\%$ overlap, and negative if they do not overlap. All other pairs are omitted. For each anchor patch $\mathcal{G}^{\mathcal{P}}_i \in \mathcal{A}$, we denote the set of its positive patches in $\mathcal{Q}$ as $\varepsilon^i_p$, and the set of its negative patches as $\varepsilon^i_n$. The overlap-aware circle loss on $\mathcal{P}$ is then defined as:
\begin{equation}
\label{eq:overlap-aware-circle-loss}
\mathcal{L}^{\mathcal{P}}_{oc} \tight{=} \frac{1}{\lvert\mathcal{A}\rvert} \sum_{\mathclap{\mathcal{G}^{\mathcal{P}}_i \in \mathcal{A}}} \log[1 + \sum_{\mathclap{\mathcal{G}^{\mathcal{Q}}_j \in \varepsilon^i_p}} e^{\lambda^j_i\beta^{i,j}_p (d^j_i - \Delta_p)} \cdot \sum_{\mathclap{\mathcal{G}^{\mathcal{Q}}_k \in \varepsilon^i_n}} e^{\beta^{i,k}_n (\Delta_n - d^k_i)}],
\end{equation}
where $d^j_i \hspace{1pt} {=} \hspace{1pt} \lVert \hat{\textbf{h}}{}^{\mathcal{P}}_i \hspace{1pt} {-} \hspace{1pt} \hat{\textbf{h}}{}^{\mathcal{Q}}_j \rVert_2$ is the distance in the feature space, $\lambda_i^j \tight{=} (o^j_i)^{\frac{1}{2}}$ and $o^j_i$ represents the overlap ratio between $\mathcal{G}^{\mathcal{P}}_i$ and $\mathcal{G}^{\mathcal{Q}}_j$. The positive and negative weights are computed for each sample individually with $\beta^{i,j}_p \medium{=} \gamma(d^j_i \medium{-} \Delta_p)$ and $\beta^{i,k}_n \medium{=} \gamma(\Delta_n \medium{-} d^k_i)$. The margin hyper-parameters are set to $\Delta_p \hspace{1pt} {=} \hspace{1pt} 0.1$ and $\Delta_n \hspace{1pt} {=} \hspace{1pt} 1.4$. The overlap-aware circle loss reweights the loss values on $\varepsilon^i_p$ based on the overlap ratio so that the patch pairs with higher overlap are given more importance. The same goes for the loss $\mathcal{L}^{\mathcal{Q}}_{oc}$ on $\mathcal{Q}$. And the overall loss is $\mathcal{L}_{oc} = (\mathcal{L}^{\mathcal{P}}_{oc} + \mathcal{L}^{\mathcal{Q}}_{oc}) / 2$.

\paraspace
\ptitle{Point matching loss.}
The ground-truth point correspondences are relatively sparse because they are available only for downsampled point clouds. We simply use a negative log-likelihood loss~\cite{sarlin2020superglue} on the assignment matrix $\bar{\textbf{Z}}_i$ of each superpoint correspondence. During training, we randomly sample $N_g$ ground-truth superpoint correspondences $\{\hat{\mathcal{C}}^{*}_i\}$ instead of using the predicted ones. For each $\hat{\mathcal{C}}^{*}_i$, a set of ground-truth point correspondences $\mathcal{M}_i$ is extracted with a matching radius $\tau$. The sets of unmatched points in the two patches are denoted as $\mathcal{I}_i$ and $\mathcal{J}_i$. The individual point matching loss for $\hat{\mathcal{C}}^{*}_i$ is computed as:
\begin{equation}
\mathcal{L}_{p, i} = -\sum_{\mathclap{{(x, y) \in \mathcal{M}_i}}} \log \bar{z}^i_{x, y} - \sum_{x \in \mathcal{I}_i} \log \bar{z}^i_{x, m_i+1} - \sum_{y \in \mathcal{J}_i} \log \bar{z}^i_{n_i+1, y},
\end{equation}
The final loss is computed by averaging the individual loss over all sampled superpoint matches: $\mathcal{L}_p = \frac{1}{N_g} \sum^{N_g}_{i=1} \mathcal{L}_{p, i}$.


\section{Experiments}
\label{sec:experiments}

In this section, we conduct extensive experiments to evaluate the effectiveness of our GeoTransformer. We first introduce the implementation details in the experiments in \cref{sec:exp-details}. Then, we evaluate our method and compare with previous state-of-the-art methods on indoor 3DMatch and 3DLoMatch benchmarks~\cite{zeng20173dmatch,huang2021predator} (\cref{sec:exp-indoor}), outdoor KITTI odometry benchmark~\cite{geiger2012we} (\cref{sec:exp-outdoor}), synthetic ModelNet40 benchmark~\cite{wu20153d} (\cref{sec:exp-synthetic}), and multiway Augmented ICL-NUIM benchmark~\cite{choi2015robust} (\cref{sec:exp-multiway}). We further investigate the generality of GeoTransformer to non-rigid registration~\cite{li2022lepard} (\cref{sec:exp-nonrigid}). Next, the ablation study is shown in \cref{sec:exp-ablation} to provide a comprehensive understanding of our design. At last, we compare our method with recent deep robust estimators in \cref{sec:exp-estimator}.

\subsection{Implementation Details}
\label{sec:exp-details}

\ptitle{Network architecture.}
As the point clouds from different benchmarks differ in density and size, we use slightly different backbones in the experiments. To be specific, we use a $4$-stage backbone for 3DMatch, ModelNet40 and 4DMatch, while a $5$-stage backbone is used for KITTI due to the much larger point clouds. Please refer to our code for more details.

In the superpoint matching module, we interleave the geometric self-attention module and the feature-based cross-attention module for $N_t \tight{=} 3$ times on all benchmarks. All attention modules have $4$ attention heads. To compute the geometric structure embedding, we simply set $\sigma_d$ to the voxel size in the superpoint level for the pair-wise distance embedding, and use $\sigma_a \tight{=} 15^{\circ}$ and $k \tight{=} 3$ for the triplet-wise angular embedding. We study the influence of these hyper-parameters in \cref{sec:exp-ablation}.

In the local-to-global registration, only the superpoint matches with at least $3$ local point correspondences are used to compute the transformation candidates. At last, we iteratively recompute the transformation with the surviving inlier matches for $N_r \tight{=} 5$ times.

\paraspace
\ptitle{Training and testing.}
We implement and evaluate GeoTransformer with PyTorch~\cite{paszke2019pytorch} on a RTX 3090 GPU. The models are trained with Adam optimizer~\cite{kingma2014adam} for $40$ epochs on 3DMatch/4DMatch, $200$ epochs on ModelNet40 and $80$ epochs on KITTI. The batch size is $1$ and the weight decay is $10^{-6}$. The learning rate starts from $10^{-4}$ and decays exponentially by $0.05$ every epoch on 3DMatch/4DMatch, every $5$ epochs on ModelNet40, and every $4$ epochs on KITTI. The same data augmentation as in~\cite{huang2021predator} is adopted. Unless otherwise noted, we randomly sample $N_g \tight{=} 128$ ground-truth superpoint correspondences during training, and use $N_c \tight{=} 256$ putative superpoint matches during testing.

\subsection{Indoor Benchmark: 3DMatch \& 3DLoMatch}
\label{sec:exp-indoor}

\ptitle{Dataset.}
3DMatch~\cite{zeng20173dmatch} contains $62$ scenes among which $46$ are used for training, $8$ for validation and $8$ for testing. We use the training data preprocessed by~\cite{huang2021predator} and evaluate on both 3DMatch and 3DLoMatch~\cite{huang2021predator} protocols. The point cloud pairs in 3DMatch have $>30\%$ overlap, while those in 3DLoMatch have low overlap of $10\%$ $\sim$ $30\%$.

\paraspace
\ptitle{Metrics.}
Following~\cite{bai2020d3feat,huang2021predator}, we evaluate the performance with three metrics: (1) \emph{Inlier Ratio} (IR), the fraction of putative correspondences whose residuals are below a certain threshold (\ie, $0.1\text{m}$) under the ground-truth transformation, (2) \emph{Feature Matching Recall} (FMR), the fraction of point cloud pairs whose inlier ratio is above a certain threshold (\ie, $5\%$), and (3) \emph{Registration Recall} (RR), the fraction of point cloud pairs whose transformation error is smaller than a certain threshold (\ie, $\mathrm{RMSE} < 0.2\text{m}$).


\begin{table}[!t]
\setlength{\tabcolsep}{2pt}
\scriptsize
\centering
\caption{
\textbf{Evaluation results on 3DMatch and 3DLoMatch}.
RANSAC is used for registration with $50$K iterations.
$^{\dagger}$ indicates the lite model with shared geometric self-attention.
\textbf{Boldfaced} numbers highlight the best and the second best are \underline{underlined}.
}
\label{table:results-3dmatch}
\begin{tabular}{l|ccccc|ccccc}
\toprule
 & \multicolumn{5}{c|}{3DMatch} & \multicolumn{5}{c}{3DLoMatch} \\
\# Samples & 5000 & 2500 & 1000 & 500 & 250 & 5000 & 2500 & 1000 & 500 & 250 \\
\midrule
\multicolumn{11}{c}{\emph{Feature Matching Recall} (\%) $\uparrow$} \\
\midrule
PerfectMatch~\cite{gojcic2019perfect} & 95.0 & 94.3 & 92.9 & 90.1 & 82.9 & 63.6 & 61.7 & 53.6 & 45.2 & 34.2 \\
FCGF~\cite{choy2019fully} & 97.4 & 97.3 & 97.0 & 96.7 & 96.6 & 76.6 & 75.4 & 74.2 & 71.7 & 67.3 \\
D3Feat~\cite{bai2020d3feat} & 95.6 & 95.4 & 94.5 & 94.1 & 93.1 & 67.3 & 66.7 & 67.0 & 66.7 & 66.5 \\
SpinNet~\cite{ao2021spinnet} & 97.6 & 97.2 & 96.8 & 95.5 & 94.3 & 75.3 & 74.9 & 72.5 & 70.0 & 63.6 \\
Predator~\cite{huang2021predator} & 96.6 & 96.6 & 96.5 & 96.3 & 96.5 & 78.6 & 77.4 & 76.3 & 75.7 & 75.3 \\
YOHO~\cite{wang2022you} & \textbf{98.2} & 97.6 & \underline{97.5} & 97.7 & 96.0 & 79.4 & 78.1 & 76.3 & 73.8 & 69.1 \\
CoFiNet~\cite{yu2021cofinet} & \underline{98.1} & \textbf{98.3} & \textbf{98.1} & \textbf{98.2} & \textbf{98.3} & 83.1& 83.5 & 83.3 & 83.1 & 82.6 \\
GeoTransformer (\emph{ours}) & \underline{98.1} & \underline{98.1} & \textbf{98.1} & \textbf{98.2} & \underline{98.1} & \underline{87.7} & \underline{87.7} & \underline{87.8} & \underline{88.0} & \underline{88.2} \\
GeoTransformer$^{\dagger}$ (\emph{ours}) & \underline{98.1} & \underline{98.1} & \textbf{98.1} & \underline{98.1} & 97.8 & \textbf{88.7} & \textbf{88.8} & \textbf{88.7} & \textbf{89.1} & \textbf{88.7} \\
\midrule
\multicolumn{11}{c}{\emph{Inlier Ratio} (\%) $\uparrow$} \\
\midrule
PerfectMatch~\cite{gojcic2019perfect} & 36.0 & 32.5 & 26.4 & 21.5 & 16.4 & 11.4 & 10.1 & 8.0 & 6.4 & 4.8 \\
FCGF~\cite{choy2019fully} & 56.8 & 54.1 & 48.7 & 42.5 & 34.1 & 21.4 & 20.0 & 17.2 & 14.8 & 11.6 \\
D3Feat~\cite{bai2020d3feat} & 39.0 & 38.8 & 40.4 & 41.5 & 41.8 & 13.2 & 13.1 & 14.0 & 14.6 & 15.0 \\
SpinNet~\cite{ao2021spinnet} & 47.5 & 44.7 & 39.4 & 33.9 & 27.6 & 20.5 & 19.0 & 16.3 & 13.8 & 11.1 \\
Predator~\cite{huang2021predator} & 58.0 & 58.4 & 57.1 & 54.1 & 49.3 & 26.7 & 28.1 & \underline{28.3} & 27.5 & 25.8 \\
YOHO~\cite{wang2022you} & 64.4 & 60.7 & 55.7 & 46.4 & 41.2 & 25.9 & 23.3 & 22.6 & 18.2 & 15.0 \\
CoFiNet~\cite{yu2021cofinet} & 49.8 & 51.2 & 51.9 & 52.2 & 52.2 & 24.4 & 25.9 & 26.7 & 26.8 & 26.9 \\
GeoTransformer (\emph{ours}) & \underline{72.5} & \underline{75.9} & \underline{76.8} & \underline{82.8} & \underline{85.6} & \textbf{44.7} & \underline{45.8} & \textbf{46.7} & \underline{53.3} & \underline{58.0} \\
GeoTransformer$^{\dagger}$ (\emph{ours}) & \textbf{72.7} & \textbf{76.1} & \textbf{76.9} & \textbf{82.9} & \textbf{85.7} & \underline{43.9} & \textbf{45.9} & \textbf{46.7} & \textbf{53.6} & \textbf{58.3} \\
\midrule
\multicolumn{11}{c}{\emph{Registration Recall} (\%) $\uparrow$} \\
\midrule
PerfectMatch~\cite{gojcic2019perfect} & 78.4 & 76.2 & 71.4 & 67.6 & 50.8 & 33.0 & 29.0 & 23.3 & 17.0 & 11.0 \\
FCGF~\cite{choy2019fully} & 85.1 & 84.7 & 83.3 & 81.6 & 71.4 & 40.1 & 41.7 & 38.2 & 35.4 & 26.8  \\
D3Feat~\cite{bai2020d3feat} & 81.6 & 84.5 & 83.4 & 82.4 & 77.9 & 37.2 & 42.7 & 46.9 & 43.8 & 39.1 \\
SpinNet~\cite{ao2021spinnet} & 88.6 & 86.6 & 85.5 & 83.5 & 70.2 & 59.8 & 54.9 & 48.3 & 39.8 & 26.8 \\
Predator~\cite{huang2021predator} & 89.0 & 89.9 & 90.6 & 88.5 & 86.6 & 59.8 & 61.2 & 62.4 & 60.8 & 58.1 \\
YOHO~\cite{wang2022you} & 90.8 & 90.3 & 89.1 & 88.6 & 84.5 & 65.2 & 65.5 & 63.2 & 56.5 & 48.0 \\
CoFiNet~\cite{yu2021cofinet} & 89.3 & 88.9 & 88.4 & 87.4 & 87.0 & 67.5 & 66.2 & 64.2 & 63.1 & 61.0 \\
GeoTransformer (\emph{ours}) & \textbf{92.3} & \textbf{92.1} & \textbf{92.0} & \textbf{91.7} & \textbf{91.2} & \textbf{75.4} & \textbf{75.0} & \textbf{74.6} & \textbf{74.0} & \textbf{73.9} \\
GeoTransformer$^{\dagger}$ (\emph{ours}) & \underline{92.2} & \underline{92.0} & \underline{91.6} & \underline{91.5} & \underline{91.1} & \underline{74.9} & \underline{74.5} & \underline{73.9} & \underline{73.6} & \underline{73.0} \\
\bottomrule
\end{tabular}
\end{table}

\paraspace
\ptitle{Correspondence results.}
We first compare the correspondence results of our method with the recent state of the arts: PerfectMatch~\cite{gojcic2019perfect}, FCGF~\cite{choy2019fully}, D3Feat~\cite{bai2020d3feat}, SpinNet~\cite{ao2021spinnet}, Predator~\cite{huang2021predator}, YOHO~\cite{wang2022you} and CoFiNet~\cite{yu2021cofinet} in \cref{table:results-3dmatch}(top and middle)\footnote{We refine our code and retrain the models, so the results are slightly better than our conference version~\cite{qin2022geometric}.}. Following~\cite{bai2020d3feat,huang2021predator}, we report the results with different numbers of correspondences. To control the number of the correspondences, we vary the hyper-parameter $k$ of the mutual top-$k$ selection in the point matching module and select the correspondences with the highest confidence scores. For \emph{Feature Matching Recall}, our method achieves improvements of at least $5$ percentage points (pp) on 3DLoMatch, demonstrating its effectiveness in low-overlap cases. For \emph{Inlier Ratio}, the improvements are even more prominent. It surpasses the baselines consistently by $8{\sim}33$ pp on 3DMatch and $18{\sim}31$ pp on 3DLoMatch. The gain is larger with less correspondences. It implies that our method extracts more reliable correspondences.

\paraspace
\ptitle{Registration results.}
\label{sec:exp-registration}
To evaluate the registration performance, we first compare the \emph{Registration Recall} obtained by RANSAC in \cref{table:results-3dmatch}(bottom). Following~\cite{bai2020d3feat,huang2021predator}, we run $50$K RANSAC iterations to estimate the transformation. GeoTransformer attains new state-of-the-art results on both 3DMatch and 3DLoMatch. It outperforms the previous best by $1.5$ pp on 3DMatch and $7.9$ pp on 3DLoMatch, showing its efficacy in both high- and low-overlap scenarios. And the shared geometric self-attention based model (\ie, the \emph{lite model}) attains very close performance to the standard one. More importantly, our method is quite stable under different numbers of samples, so it does not require sampling a large number of correspondences to boost the performance as previous methods \cite{choy2019fully,ao2021spinnet,wang2022you,yu2021cofinet}.


\begin{table}[!t]
\scriptsize
\setlength{\tabcolsep}{1pt}
\centering
\caption{
\textbf{Registration results w/o RANSAC} on 3DMatch (3DM) and 3DLoMatch (3DLM).
The \emph{model time} is the time for feature extraction, while the \emph{pose time} is the time for transformation estimation.
The time is averaged over all point cloud pairs in 3DMatch and 3DLoMatch.
$^{\dagger}$ indicates the lite model with shared geometric self-attention.
\textbf{Boldfaced} numbers highlight the best and the second best are \underline{underlined}.
}
\label{table:direct-3dmatch}
\begin{tabular}{l|c|c|cc|ccc}
\toprule
\multirow{2}{*}{Model} & \multirow{2}{*}{Estimator} & \multirow{2}{*}{\#Samples} & \multicolumn{2}{c|}{RR(\%)} & \multicolumn{3}{c}{Time(s)} \\
 & & & 3DM & 3DLM & Model & Pose & Total\\
\midrule
FCGF~\cite{choy2019fully} & RANSAC-\emph{50k} & 5000 & 85.1 & 40.1 & 0.052 & 3.326 & 3.378 \\
D3Feat~\cite{bai2020d3feat} & RANSAC-\emph{50k} & 5000 & 81.6 & 37.2 & 0.024 & 3.088 & 3.112 \\
SpinNet~\cite{ao2021spinnet} & RANSAC-\emph{50k} & 5000 & 88.6 & 59.8 & 60.248 & 0.388 & 60.636 \\
Predator~\cite{huang2021predator} & RANSAC-\emph{50k} & 5000 & 89.0 & 59.8 & 0.032 & 5.120 & 5.152 \\
CoFiNet~\cite{yu2021cofinet} & RANSAC-\emph{50k} & 5000 & 89.3 & 67.5 & 0.115 & 1.807 & 1.922 \\
GeoTransformer (\emph{ours}) & RANSAC-\emph{50k} & 5000 & \textbf{92.3} & \textbf{75.4} & 0.075 & 1.558 & 1.633 \\
GeoTransformer$^{\dagger}$ (\emph{ours}) & RANSAC-\emph{50k} & 5000 & \underline{92.2} & \underline{74.9} & 0.060 & 1.546 & 1.606 \\
\midrule
FCGF~\cite{choy2019fully} & weighted SVD & 250 & 42.1 & 3.9 & 0.052 & 0.008 & 0.056 \\  
D3Feat~\cite{bai2020d3feat} & weighted SVD & 250 & 37.4 & 2.8 & 0.024 & 0.008 & 0.032 \\  
SpinNet~\cite{ao2021spinnet} & weighted SVD & 250 & 34.0 & 2.5 & 60.248 & 0.006 & 60.254 \\  
Predator~\cite{huang2021predator} & weighted SVD & 250 & 50.0 & 6.4 & 0.032 & 0.009 & 0.041 \\  
CoFiNet~\cite{yu2021cofinet} & weighted SVD & 250 & 64.6 & 21.6 & 0.115 & 0.003 & 0.118 \\
GeoTransformer (\emph{ours}) & weighted SVD & 250 & \underline{86.7} & \underline{60.5} & 0.075 & 0.003 & 0.078 \\
GeoTransformer$^{\dagger}$ (\emph{ours}) & weighted SVD & 250 & \textbf{87.5} & \textbf{61.4} & 0.060 & 0.003 & 0.063 \\
\midrule
CoFiNet~\cite{yu2021cofinet} & LGR & all & 87.6 & 64.8 & 0.115 & 0.028 & 0.143 \\
GeoTransformer (\emph{ours}) & LGR & all & \textbf{91.8} & \textbf{74.5} & 0.075 & 0.013 & 0.088 \\
GeoTransformer$^{\dagger}$ (\emph{ours}) & LGR & all & \textbf{91.8} & \underline{74.2} & 0.060 & 0.013 & 0.073 \\
\bottomrule
\end{tabular}
\end{table}


\begin{figure*}[t]
  \centering
  \begin{overpic}[width=1.0\linewidth]{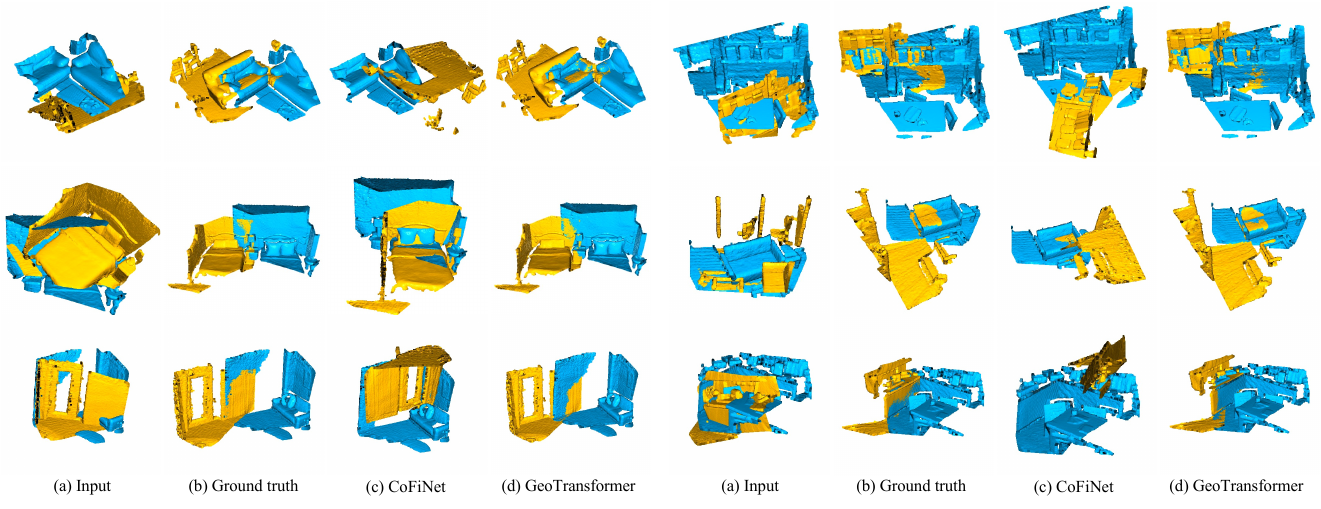}
  \end{overpic}
  \caption{\textbf{Comparison of the registration results on 3DLoMatch}. GeoTransformer can effectively recognize small overlapping area in complex scenes (see the first and third rows on the right) and distinguish similar objects at different positions (see the second and third rows on the left) thanks to the structure information from geometric self-attention.
  }
  \label{fig:gallery-3dmatch}
\end{figure*}

We then compare the registration results \emph{without} using RANSAC in \cref{table:direct-3dmatch}. We start with weighted SVD over correspondences in solving for alignment transformation. For the baselines, we first sample $5000$ keypoints and generate the correspondences with mutual nearest neighbor selection on their descriptors, and then the top $250$ correspondences are used to compute the transformation. The baselines either fail to achieve reasonable results or suffer from severe performance degradation. In contrast, GeoTransformer (with weighted SVD) achieves the registration recall of $86.7\%$ on 3DMatch and $60.5\%$ on 3DLoMatch, close to Predator with RANSAC. Note that the lite model performs even better than the standard model thanks to the higher inlier ratio. Without outlier filtering by RANSAC, high inlier ratio is necessary for successful registration. However, high inlier ratio does not necessarily lead to high registration recall since the correspondences could cluster together as noted in~\cite{huang2021predator}. Nevertheless, our method without RANSAC performs well by extracting reliable and well-distributed superpoint correspondences.

When using our local-to-global registration (LGR) for computing transformation, our method brings the registration recall to $91.8\%$ on 3DMatch and $74.5\%$ on 3DLoMatch, surpassing all RANSAC-based baselines by a large margin. The results are also very close to those of ours with RANSAC, but LGR gains over $100$ times acceleration over RANSAC in the pose time. These results demonstrate the superiority of our method in both accuracy and speed. Moreover, our lite model achieves very similar results but reduces the overall time by $17\%$, running at $13$ fps. We believe it has a good potential in real-time applications such as online 3D reconstruction and camera relocalization.


\begin{figure}[t]
  \begin{overpic}[width=1.0\linewidth]{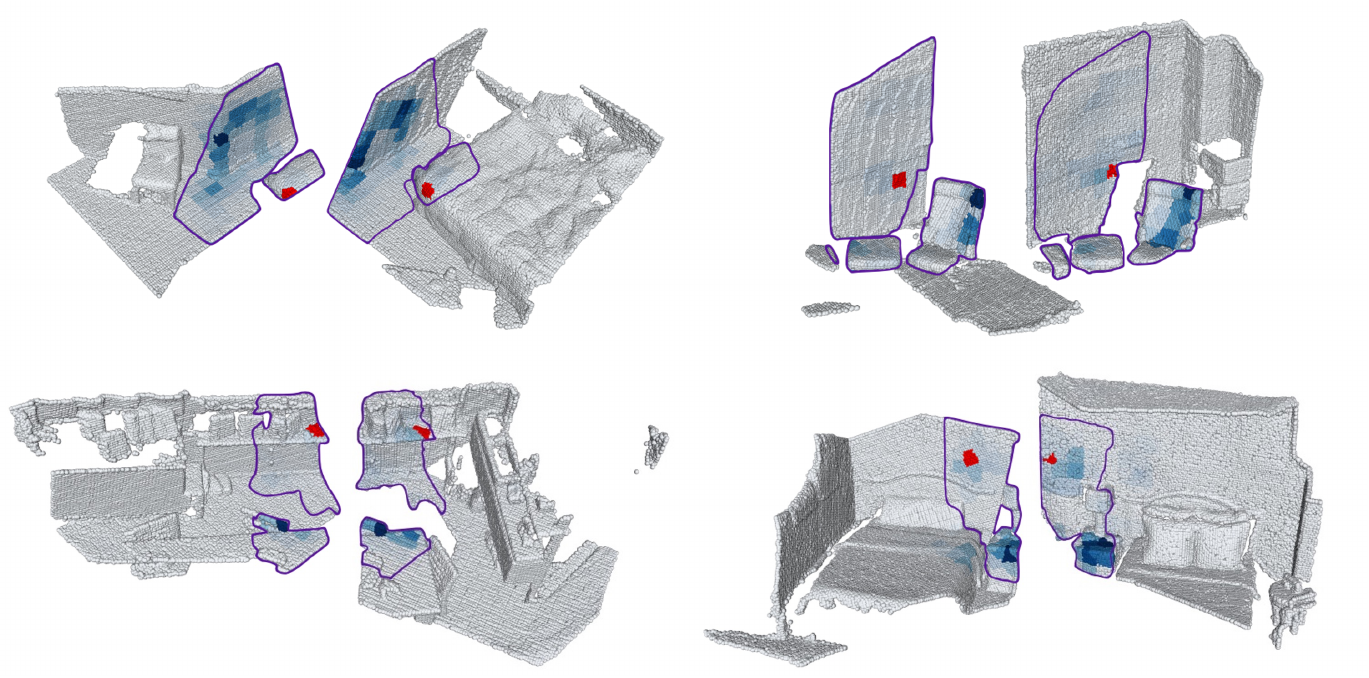}
  \end{overpic}
  \caption{
  \textbf{Visualizing geometric self-attention scores} on four pairs of point clouds.
  The overlap areas are delineated with purple lines.
  The anchor patches (in correspondence) are highlighted in red and the attention scores to other patches are color-coded (\emph{deeper is larger}).
  Note how the attention patterns of the two matching anchors are consistent even across disjoint overlap areas.
  }
  \label{fig:attention}
\end{figure}

\paraspace
\ptitle{Qualitative results.}
We provide some qualitative comparison of CoFiNet~\cite{yu2021cofinet} and GeoTransformer on 3DLoMatch in \cref{fig:gallery-3dmatch}. Our method performs quite well in these low-overlap cases. It is noteworthy that our method can distinguish similar objects at different positions (see the comparison of CoFiNet and GeoTransformer in the $2^{\text{nd}}$ and $3^{\text{rd}}$ rows on the left) and recognize small overlapping regions in complex environment thanks to the geometric structure information obtained from the geometric self-attention.

\cref{fig:attention} visualizes the attention scores learned by our geometric self-attention, which exhibits significant consistency between the anchor patch matches. It shows that our method is able to learn inter-point-cloud geometric consistency which is important to accurate correspondences.

\subsection{Outdoor Benchmark: KITTI odometry}
\label{sec:exp-outdoor}

\ptitle{Dataset.}
KITTI odometry~\cite{geiger2012we} consists of 11 sequences of outdoor driving scenarios scanned by LiDAR. We follow~\cite{choy2019fully,bai2020d3feat,huang2021predator} and use sequences 0-5 for training, 6-7 for validation and 8-10 for testing. As in~\cite{choy2019fully,bai2020d3feat,huang2021predator}, the ground-truth poses are refined with ICP and we only use point cloud pairs that are at least $10\text{m}$ away for evaluation.

\paraspace
\ptitle{Metrics.}
We follow~\cite{huang2021predator} to evaluate our GeoTransformer with three metrics: (1) \emph{Relative Rotation Error} (RRE), the geodesic distance between estimated and ground-truth rotation matrices, (2) \emph{Relative Translation Error} (RTE), the Euclidean distance between estimated and ground-truth translation vectors, and (3) \emph{Registration Recall} (RR), the fraction of point cloud pairs whose RRE and RTE are both below certain thresholds (\ie, RRE$<$5$^\circ$ and RTE$<$2m).

\paraspace
\ptitle{Registration results.}
In \cref{table:results-kitti}(top), we compare to the state-of-the-art \emph{RANSAC-based} methods: 3DFeat-Net~\cite{yew20183dfeat}, FCGF~\cite{choy2019fully}, D3Feat~\cite{bai2020d3feat}, SpinNet~\cite{ao2021spinnet}, Predator~\cite{huang2021predator} and CoFiNet~\cite{yu2021cofinet}. Our method performs on par with these methods, showing good generality on outdoor scenes. Note that the backbone in Predator is $2$ times wider than that in GeoTransformer, demonstrating the efficacy and the parameter efficiency of our method.

We further compare to three \emph{RANSAC-free} methods in \cref{table:results-kitti}(bottom): FMR~\cite{huang2020feature}, DGR~\cite{choy2020deep} and HRegNet~\cite{lu2021hregnet}. Our method outperforms all the baselines by large margin. In addition, our method with LGR beats all the RANSAC-based methods. To the best of our knowledge, GeoTransformer is the \emph{first} RANSAC-free method that surpasses RANSAC-based methods on this benchmark.


\begin{table}[!t]
\scriptsize
\centering
\caption{
\textbf{Registration results on KITTI odometry}.
\textbf{Boldfaced} numbers highlight the best and the second best are \underline{underlined}.
}
\label{table:results-kitti}
\begin{tabular}{l|ccc}
\toprule
Model & RTE(cm) & RRE($^{\circ}$) & RR(\%) \\
\midrule
3DFeat-Net~\cite{yew20183dfeat} & 25.9 & \textbf{0.25} & 96.0 \\
FCGF~\cite{choy2019fully} & 9.5 & 0.30 & \underline{96.6} \\
D3Feat~\cite{bai2020d3feat} & \underline{7.2} & 0.30 & \textbf{99.8} \\
SpinNet~\cite{ao2021spinnet} & 9.9 & 0.47 & 99.1 \\
Predator~\cite{huang2021predator} & \textbf{6.8} & \underline{0.27} & \textbf{99.8} \\
CoFiNet~\cite{yu2021cofinet} & 8.2 & 0.41 & \textbf{99.8} \\
GeoTransformer (\emph{ours}, RANSAC-\emph{50k}) & 7.4 & \underline{0.27} & \textbf{99.8} \\
\midrule
FMR~\cite{huang2020feature} & $\sim$66 & 1.49 & 90.6 \\
DGR~\cite{choy2020deep} & $\sim$32 & 0.37 & 98.7 \\
HRegNet~\cite{lu2021hregnet} & $\sim$\underline{12} & \underline{0.29} & \underline{99.7} \\
GeoTransformer (\emph{ours}, LGR) & \textbf{6.8} & \textbf{0.24} & \textbf{99.8} \\
\bottomrule
\end{tabular}
\end{table}

\subsection{Synthetic Benchmark: ModelNet40}
\label{sec:exp-synthetic}

\ptitle{Dataset.}
ModelNet40~\cite{wu20153d} contains man-made CAD models from 40 categories. Following~\cite{yew2020rpm,huang2021predator}, we use the processed data from~\cite{qi2017pointnet}, which uniformly samples $2048$ points on the surface of each CAD model. We first normalize the CAD model into a unit sphere and adopt the same strategy as in~\cite{yew2020rpm} to generate the source and the target point clouds: a half-space with a random direction is sampled and shifted to retain a proportion $p$ of the points. The source point cloud is then randomly transformed with a rotation within $[0, r]$ and a translation within $[-0.5, 0.5]$. Both point clouds are then jittered with a noise sampled from $\mathcal{N}(0, 0.01)$ and clipped to $[-0.05, 0.05]$. At last, $717$ points are randomly sampled from each point cloud independently as the final point cloud pair. We evaluate our method on two overlap settings (\emph{ModelNet} with $p \tight{=} 0.7$ and \emph{ModelLoNet} with $p \tight{=} 0.5$) and two rotation settings (\emph{Small} with $r \tight{=} 45^{\circ}$ and \emph{Large} with $r \tight{=} 180^{\circ}$). We follow~\cite{huang2021predator} to use the first $20$ categories in the official training/testing split for training/validation and the other $20$ categories in the official testing split for testing. We further remove $8$ symmetric categories (\ie, bottle, bowl, cone, cup, flower pot, lamp, tent, and vase) as their poses are ambiguous. As a result, we have $4194$ CAD models for training, $1002$ for validation, and $1146$ for testing.

\paraspace
\ptitle{Metrics.}
We follow~\cite{yew2020rpm} to evaluate GeoTransformer with two metrics: (1) \emph{Relative Rotation Error} (RRE), (2) \emph{Relative Translation Error} (RTE), and (3) \emph{Chamfer Distance} (CD) between two aligned point clouds. And we use the modified Chamfer distance from~\cite{yew2020rpm} which compares with the clean and complete versions of the other point cloud.


\begin{table}[!t]
\scriptsize
\setlength{\tabcolsep}{4pt}
\centering
\caption{
\textbf{Registration results on ModelNet40}.
\textbf{Boldfaced} numbers hightlight the best and the second best are \underline{underlined}.
}
\label{table:results-modelnet}
\begin{tabular}{l|ccc|ccc}
\toprule
\multirow{2}{*}{Model} & \multicolumn{3}{c|}{ModelNet} & \multicolumn{3}{c}{ModelLoNet} \\
 & RRE($^{\circ}$) & RTE & CD & RRE($^{\circ}$) & RTE & CD \\
\midrule
\multicolumn{7}{c}{Small Rotation} \\
\midrule
RPM-Net~\cite{yew2020rpm} & 2.357 & 0.028 & \textbf{0.00130} & 8.123 & 0.086 & \underline{0.00611} \\
RGM~\cite{fu2021robust} & 4.548 & 0.049 & 0.00268 & 14.806 & 0.139 & 0.01482 \\
Predator~\cite{huang2021predator} & \textbf{2.064} & \textbf{0.023} & 0.00145 & \underline{5.022} & \underline{0.084} & 0.00734 \\
CoFiNet~\cite{yu2021cofinet} & 3.584 & 0.044 & 0.00205 & 6.992 & 0.091 & 0.00599 \\
GeoTransformer (\emph{ours}) & \underline{2.160} & \underline{0.024} & \underline{0.00143} & \textbf{3.638} & \textbf{0.064} & \textbf{0.00448} \\
\midrule
\multicolumn{7}{c}{Large Rotation} \\
\midrule
RPM-Net~\cite{yew2020rpm} & 31.509 & 0.206 & 0.01074 & 51.478 & 0.346 & \underline{0.01985} \\
RGM~\cite{fu2021robust} & 45.560 & 0.289 & 0.01697 & 68.724 & 0.442 & 0.03634 \\
Predator~\cite{huang2021predator} & 24.839 & 0.171 & 0.01940 & 46.990 & 0.378 & 0.05052 \\
CoFiNet~\cite{yu2021cofinet} & \underline{10.496} & \underline{0.084} & \underline{0.00319} & \underline{32.578} & \underline{0.226} & 0.02273 \\
GeoTransformer (\emph{ours}) & \textbf{6.436} & \textbf{0.047} & \textbf{0.00154} & \textbf{23.478} & \textbf{0.152} & \textbf{0.01296} \\
\bottomrule
\end{tabular}
\end{table}

\paraspace
\ptitle{Registration results.}
We compare GeoTransformer with four baseline methods in \cref{table:results-modelnet}: RPM-Net~\cite{yew2020rpm}, RGM~\cite{fu2021robust}, Predator~\cite{huang2021predator} and CoFiNet~\cite{yu2021cofinet}. RPM-Net and RGM are end-to-end registration methods, while Predator, CoFiNet and GeoTransformer are correspondence-based methods. All the models are train for $200$ epochs. For fair comparison, we adopt the same KPConv-based backbone in Predator, CoFiNet and GeoTransformer. To estimate the transformation, Predator and CoFiNet use RANSAC-$50k$ while LGR is used in GeoTransformer. As the point clouds are relatively small, we use $N_c \tight{=} 128$ superpoint correspondences during testing.

When the rotation is small, RPM-Net, Predator and GeoTransformer achieve comparable results on the high-overlap setting. As this setting is relatively easy, the performance tends to be saturated. For the low-overlap setting, GeoTransformer surpasses other methods by a large margin, demonstrating the effectiveness of our method. When the rotation is large, as all the methods are not completely invariant to transformation (\ie, the backbone part), the performance inevitably drops compared with the small-rotation setting. Nevertheless, as the geometric self-attention provides more structure information about the point clouds, our GeoTransformer attains significantly better results than the baseline methods on both high- and low-overlap settings, showing strong robustness to low overlap and large rotations. Moreover, it is noteworthy that GeoTransformer requires neither iterative registration as in RPM-Net and RGM, nor RANSAC as in Predator and CoFiNet, thus achieves very fast registration speed.


\begin{table}[!t]
\scriptsize
\setlength{\tabcolsep}{5pt}
\centering
\caption{
\textbf{Registration results on Augmented ICL-NUIM}. ATE (cm) are reported.
\textbf{Boldfaced} numbers highlight the best and the second best are \underline{underlined}.
}
\label{table:results-multiway}
\begin{tabular}{l|ccccc}
\toprule
Model & Living1 & Living2 & Office1 & Office2 & Mean \\
\midrule
FGR~\cite{zhou2016fast} & 78.97 & 24.91 & 14.96 & 21.05 & 34.98 \\
RANSAC~\cite{fischler1981random} & 110.9 & 19.33 & 14.42 & 17.31 & 40.49 \\
DGR~\cite{choy2020deep} & 21.06 & 21.88 & 15.76 & 11.56 & 17.57 \\
PointDSC~\cite{bai2021pointdsc} & \underline{20.25} & \underline{15.58} & \textbf{13.56} & \underline{11.30} & \underline{15.18} \\
GeoTransformer (\emph{ours}) & \textbf{17.54} & \textbf{15.31} & \underline{13.85} & \textbf{9.78} & \textbf{14.12} \\
\bottomrule
\end{tabular}
\end{table}

Besides, we have more interesting observations. First, the correspondence-based methods perform much better than the end-to-end methods in complicated scenarios with large perturbations or heavy occlusion. In this case, the end-to-end methods have difficulty learning accurate soft correspondences, while the correspondence-based methods are more robust because they establish hard correspondences directly from the existing points. And robust estimators such as RANSAC further improve the stability of them. Second, the coarse-to-fine methods are more robust to large rotations than the detection-based methods. Compared with directly matching dense points with their descriptors, the superpoint matching is more sparse and discriminative. The two-stage pipeline can effectively alleviate the risk of mismatching and contributes to better registration performance.

\subsection{Multiway Benchmark: Augmented ICL-NUIM}
\label{sec:exp-multiway}

\ptitle{Dataset.}
Augmented ICL-NUIM~\cite{choi2015robust} augments the synthetic scenes in ICL-NUIM~\cite{handa2014benchmark} with a realistic noise model. It consists of four camera trajectories from two scenes for testing. Following~\cite{choy2020deep,bai2021pointdsc}, we fuse $50$ consecutive RGB-D frames to generate the point cloud fragements. To solve multiway registration, we follow~\cite{choy2020deep,bai2021pointdsc} to first conduct pair-wise registration with GeoTransformer and then optimize the poses with the global pose graph optimization~\cite{kummerle2011g} implemented in~\cite{zhou2018open3d}.


\begin{figure}[t]
  \centering
  \begin{overpic}[width=1.0\linewidth]{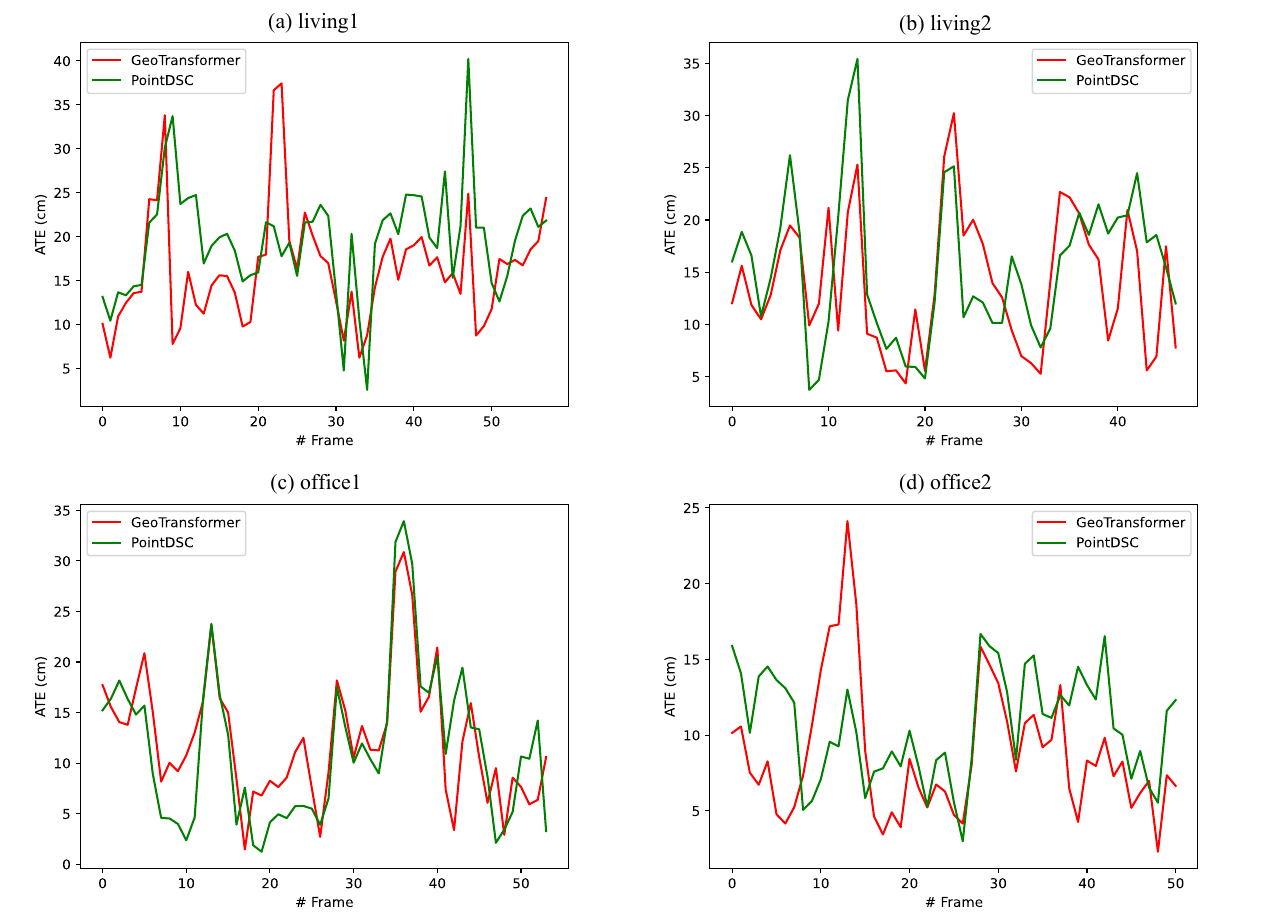}
  \end{overpic}
  \caption{\textbf{Comparison of per-frame ATE on Augmented ICL-NUIM}. Our GeoTransformer attains better results on most of the frames.
  }
  \label{fig:gallery-redwood}
\end{figure}

\paraspace
\ptitle{Metrics.}
We follow~\cite{choy2020deep,bai2021pointdsc} to evaluate GeoTransformer with the metric of \emph{Absolute Trajectory Error} (ATE). It first aligns the ground-truth and the estimated trajectories with SVD, and then computes the root mean square error (RMSE) of the differences between the points at the same timestamp in the two trajectories.

\paraspace
\ptitle{Registration results.}
We compare GeoTransformer with four baseline methods in \cref{table:results-multiway}: FGR~\cite{zhou2016fast}, RANSAC~\cite{fischler1981random}, DGR~\cite{choy2020deep}, and PointDSC~\cite{bai2021pointdsc}. Following~\cite{bai2021pointdsc}, we directly use the models trained on 3DMatch without fine-tuning. However, the point clouds in Augmented ICL-NUIM are larger than those in 3DMatch due to faster camera motion, so we further downsample the superpoints to reduce memory footprint. And the superpoint features $\hat{\mathbf{F}}^{\mathcal{P}}$ and $\hat{\mathbf{F}}^{\mathcal{Q}}$ are downsampled by $k$NN interpolation, where the geometric transformer module is then applied. At last, the resultant features are interpolated and upsampled to generate $\hat{\mathbf{H}}^{\mathcal{P}}$ and $\hat{\mathbf{H}}^{\mathcal{Q}}$. This modification effectively improves the memory efficiency without sacrificing the performance. GeoTransformer attains the best performance on all testing trajectories except \texttt{Office1}, showing strong generality to unknown scenes and more complex applications.
\cref{fig:gallery-redwood} visualizes the ATE of each frame in the four trajectories and GeoTransformer achieves better results on most of the frames.

\subsection{Non-rigid Benchmark: 4DMatch \& 4DLoMatch}
\label{sec:exp-nonrigid}


\begin{figure*}[t]
  \centering
  \begin{overpic}[width=1.0\linewidth]{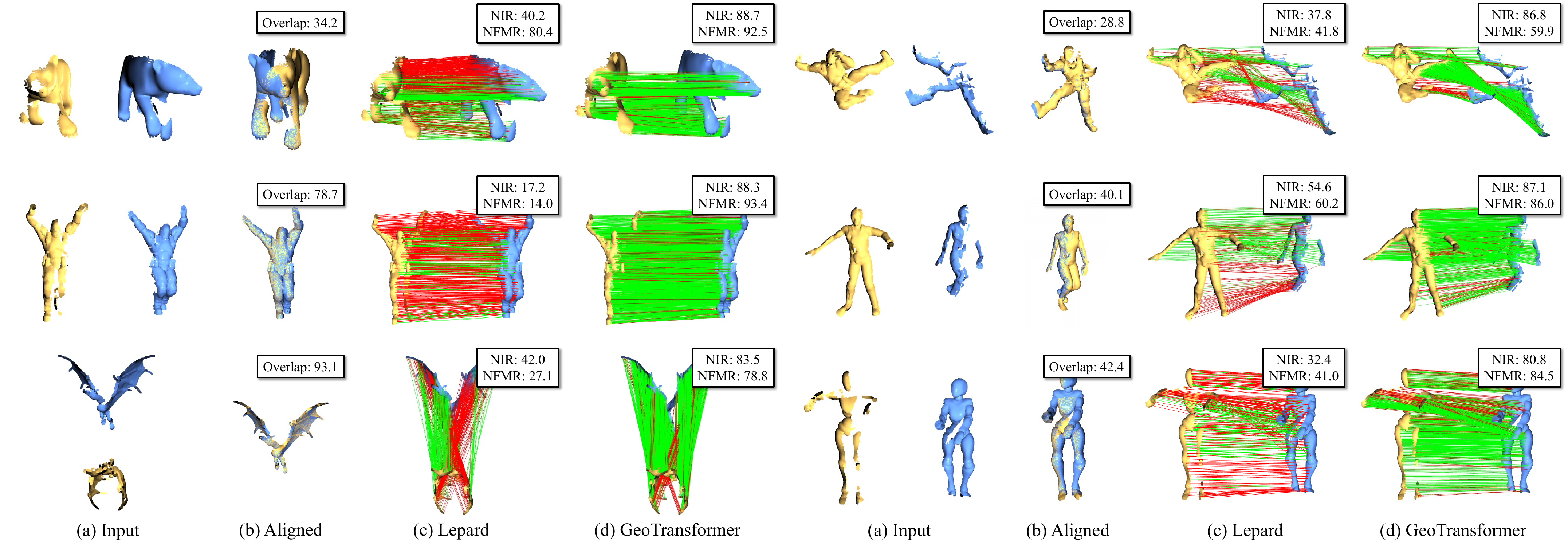}
  \end{overpic}
  \caption{\textbf{Comparison of the correspondences on 4DMatch and 4DLoMatch}. GeoTransformer shows two advantages. First, it extracts much denser correspondences, which contributes to more precise description of the deformations. Second, it achieves higher inlier ratio despite significant deformations, which is important for non-rigid registration.
  }
  \label{fig:gallery-4dmatch}
\end{figure*}

\ptitle{Dataset.}
4DMatch~\cite{li2022lepard} is a challenging benchmark for non-rigid point cloud registration. It is constructed using the animation sequences from DeformingThings4D~\cite{li20214dcomplete}, where $1232$ sequences are used for training, $176$ for validation and $353$ for testing. The point cloud pairs in the testing sequences are divided into 4DMatch and 4DLoMatch based on an overlapping ratio threshold of $45\%$.

\paraspace
\ptitle{Metrics.}
Following~\cite{li2022lepard}, we evaluate our GeoTransformer with two metrics: (1) \emph{Non-rigid Inlier Ratio} (NIR), the fraction of putative correspondences whose residuals are below a certain threshold (\ie, $0.04\text{m}$) under the ground-truth warping function, and (2) \emph{Non-rigid Feature Matching Recall} (NFMR), the fraction of the ground-truth matches that can be successfully recovered by the putative correspondences.

\paraspace
\ptitle{Implementation details.}
Unlike rigid registration which can be pinned down by a set of sparse correspondences, non-rigid registration is more challenging  due to the complex and irregular deformation and requires denser correspondences to cover the overlapping region as much as possible. To this end, we modify our superpoint matching strategy to increase the overall coverage of the correspondences. Specifically, we first select the superpoint matches whose feature distances are below a certain threshold (\ie, 0.75), and augment them with the top $128$ ones if there are too few superpoint matches. We use all point correspondences for evaluation and LGR is not performed.

\paraspace
\ptitle{Evaluation results.}
We compare GeoTransformer with three recent methods in \cref{table:results-4dmatch}: D3Feat~\cite{bai2020d3feat}, Predator~\cite{huang2021predator} and Lepard~\cite{li2022lepard}. Our method surpasses D3Feat and Predator by a large margin on both high- and low-overlap scenarios. Compared with Lepard, GeoTransformer achieves very close performance on 4DMatch and significantly better inlier ratio on 4DLoMatch. Note that Lepard benefits from a repositioning mechanism with a coarse rigid registration, which effectively boosts the performance. Without repositioning, our GeoTransformer consistently outperforms Lepard on both benchmarks. Albeit not carefully designed and optimized to handle deformation, GeoTransformer shows strong generality to non-rigid registration. In most cases, the non-rigid deformation could be approximated by a set of local rigid transformations. We suppose that the novel geometric self-attention emdows GeoTransformer with the capability to capture the local rigidity consistency between two point clouds, which helps extracting high-quality correspondences in non-rigid scenarios.


\begin{table}[!t]
\scriptsize
\setlength{\tabcolsep}{5pt}
\centering
\caption{
\textbf{Evaluation results on 4DMatch and 4DLoMatch}.
NIR and NFMR are measured in \%.
Our method uses shared geometric self-attention.
\textbf{Boldfaced} numbers are the best and the second best are \underline{underlined}.
}
\label{table:results-4dmatch}
\begin{tabular}{l|ccc|ccc}
\toprule
\multirow{2}{*}{Model} & \multicolumn{3}{c|}{4DMatch} & \multicolumn{3}{c}{4DLoMatch} \\
 & \# Corr & NIR & NFMR & \# Corr & NIR & NFMR \\
\midrule
D3Feat~\cite{bai2020d3feat} & 697 & 55.3 & 56.1 & 204 & 21.3 & 28.1 \\
Predator~\cite{huang2021predator} & 698 & 59.3 & 56.8 & 480 & 25.0 & 32.1 \\
Lepard~\cite{li2022lepard} & 596 & \textbf{82.7} & \textbf{83.7} & 407 & \underline{55.7} & \textbf{66.9} \\
Lepard (w/o repos)~\cite{li2022lepard} & 624 & 80.5 & 80.8 & 448 & 53.7 & 63.6 \\
GeoTransformer (\emph{ours}) & 2331 & \underline{82.2} & \underline{83.2} & 1212 & \textbf{63.6} & \underline{65.4} \\
\bottomrule
\end{tabular}
\end{table}

\paraspace
\ptitle{Qualitative results.}
\cref{fig:gallery-4dmatch} compares the correspondences from Lepard~\cite{li2022lepard} and GeoTransformer on some cases with relatively large deformations. GeoTransformer has two advantages as shown in these cases. First, our method can extract much denser correspondences, which is important for precisely describing the deformation. Due to the irregularity of the deformations, it is diffcult to capture the deformation details if the correspondences are too sparse. Second, our method attains much higher inlier ratio especially in low-overlap cases, which benifits the following registration algorithms such as Non-rigid ICP~\cite{li2008global,yao2020quasi}. As there are few effective outlier rejection methods for non-rigid registration, high inlier ratio is crucial for estimating a proper deformation field. Thanks to the coarse-to-fine framework and the geometric self-attention, our method can establish high-quality correspondences for non-rigid registration.

\subsection{Ablation Studies}
\label{sec:exp-ablation}

We conduct extensive ablation studies on 3DMatch and 3DLoMatch for a better understanding of the various modules in our method. To evaluate superpoint (patch) matching, we introduce another metric \emph{Patch Inlier Ratio} (PIR) which is the fraction of patch matches with actual overlap. The FMR and IR are reported with \emph{all} global dense point correspondences, with LGR being used for registration.


\begin{table}[!t]
\scriptsize
\setlength{\tabcolsep}{2.5pt}
\centering
\caption{
\textbf{Ablation experiments on 3DMatch and 3DLoMatch}.
The results are measured in \%.
* indicates the default settings of GeoTransformer.
\textbf{Boldfaced} numbers are the best and the second best are \underline{underlined}.
}
\label{table:ablation-study}
\begin{tabular}{l|cccc|cccc}
\toprule
\multirow{2}{*}{Model} & \multicolumn{4}{c|}{3DMatch} & \multicolumn{4}{c}{3DLoMatch} \\
 & PIR & FMR & IR & RR & PIR & FMR & IR & RR \\
\midrule
(a.1) graph neural network & 73.3 & 97.9 & 56.5 & 89.5 & 39.4 & 84.9 & 29.2 & 69.8 \\
(a.2) vanilla self-attention & 79.6 & 97.9 & 60.1 & 89.0 & 45.2 & 85.6 & 32.6 & 68.4 \\
(a.3) self-attention w/ ACE & 83.2 & \textbf{98.1} & 68.5 & 89.3 & 48.2 & 84.3 & 38.9 & 69.3 \\
(a.4) self-attention w/ RCE & 80.0 & 97.9 & 66.1 & 88.5 & 46.1 & 84.6 & 37.9 & 68.7 \\
(a.5) self-attention w/ PPF & 83.5 & 97.5 & 68.5 & 88.6 & 49.8 & 83.8 & 39.9 & 69.5 \\
(a.6) self-attention w/ RDE & \underline{84.9} & \underline{98.0} & \underline{69.1} & \underline{90.7} & \underline{50.6} & \underline{85.8} & \underline{40.3} & \underline{72.1} \\
(a.7) geometric self-attention* & \textbf{86.1} & \textbf{98.1} & \textbf{71.0} & \textbf{91.8} & \textbf{54.6} & \textbf{87.8} & \textbf{43.8} & \textbf{74.5} \\
\midrule
(b.1) cross-entropy loss & 80.0 & 97.7 & 65.7 & 90.0 & 45.9 & 85.1 & 37.4 & 68.4 \\
(b.2) weighted cross-entropy loss & 83.2 & \underline{98.0} & 67.4 & 90.0 & 49.0 & \underline{86.2} & 38.6 & 70.7 \\
(b.3) circle loss & \underline{85.1} & 97.8 & \underline{69.5} & \underline{90.4} & \underline{51.5} & 86.1 & \underline{41.3} & \underline{71.5} \\
(b.4) overlap-aware circle loss* & \textbf{86.1} & \textbf{98.1} & \textbf{71.0} & \textbf{91.8} & \textbf{54.6} & \textbf{87.8} & \textbf{43.8} & \textbf{74.5} \\
\midrule
(c.1) distance only & 84.9 & 98.0 & 69.1 & 90.7 & 50.6 & 85.8 & 40.3 & 72.1 \\
(c.2) $k=1$ angles & \textbf{86.5} & 97.9 & 70.6 & 91.0 & 54.6 & 87.1 & 42.7 & 73.1 \\
(c.3) $k=2$ angles & 86.1 & 97.9 & 70.4 & 91.3 & \textbf{55.0} & \textbf{88.2} & 43.5 & 73.5 \\
(c.4) $k=3$ angles* & 86.1 & \underline{98.1} & \underline{71.0} & \underline{91.8} & 54.6 & \underline{87.8} & \underline{43.8} & \underline{74.5} \\
(c.5) $k=4$ angles & \underline{86.4} & \textbf{98.2} & \textbf{71.1} & \textbf{92.1} & \underline{54.8} & \underline{87.8} & \textbf{43.9} & \textbf{75.1} \\
\midrule
(d.1) $\sigma_d=0.1\text{m}$ & \underline{86.6} & 97.6 & \textbf{71.4} & 90.7 & \underline{54.6} & \underline{87.7} & \textbf{43.8} & 73.2 \\
(d.2) $\sigma_d=0.2\text{m}$* & 86.1 & \underline{98.1} & \underline{71.0} & \underline{91.8} & \underline{54.6} & \textbf{87.8} & \textbf{43.8} & \textbf{74.5} \\
(d.3) $\sigma_d=0.3\text{m}$ & \textbf{86.7} & \textbf{98.4} & 70.3 & \textbf{92.0} & \textbf{55.3} & 87.0 & 43.0 & \underline{74.1} \\
(d.4) $\sigma_d=0.4\text{m}$ & \textbf{86.7} & \underline{98.1} & \textbf{71.4} & \underline{91.8} & 54.3 & \textbf{87.8} & \underline{43.5} & 74.0 \\
(d.5) $\sigma_d=0.5\text{m}$ & 86.0 & 97.8 & 70.3 & 91.0 & 54.2 & 86.3 & 43.3 & 73.7 \\
\midrule
(e.1) $\sigma_a=5^{\circ}$ & 86.1 & 97.9 & 70.4 & 91.3 & 53.7 & 86.9 & 42.4 & 72.6 \\
(e.2) $\sigma_a=10^{\circ}$ & \textbf{87.0} & \underline{98.0} & \textbf{71.4} & 91.4 & 54.5 & \underline{87.3} & \underline{43.6} & \underline{74.2} \\
(e.3) $\sigma_a=15^{\circ}$* & 86.1 & \textbf{98.1} & \underline{71.0} & \underline{91.8} & \underline{54.6} & \textbf{87.8} & \textbf{43.8} & \textbf{74.5} \\
(e.4) $\sigma_a=20^{\circ}$ & \underline{86.7} & 97.9 & 70.7 & \textbf{92.1} & \textbf{54.7} & 86.7 & 43.0 & 73.6 \\
(e.5) $\sigma_a=25^{\circ}$ & 86.5 & 97.8 & 70.6 & 91.2 & 54.0 & 86.6 & 42.7 & 73.6 \\
\midrule
(f.1) w/ max pooling* & 86.1 & \textbf{98.1} & \textbf{71.0} & \textbf{91.8} & \textbf{54.6} & \textbf{87.8} & \textbf{43.8} & \textbf{74.5} \\
(f.2) w/ average pooling & \textbf{86.3} & 98.0 & 70.2 & 91.3 & \textbf{54.6} & 87.3 & 42.8 & 74.0 \\
\midrule
(g.1) w/ dual-normalization* & 86.1 & \textbf{98.1} & \textbf{71.0} & \textbf{91.8} & \textbf{54.6} & 87.8 & \textbf{43.8} & \textbf{74.5} \\
(g.2) w/o dual-normalization & \textbf{86.2} & \textbf{98.1} & 70.9 & \textbf{91.8} & 53.5 & \textbf{87.9} & 43.4 & 74.4 \\
\bottomrule
\end{tabular}
\end{table}


\begin{figure*}[t]
  \begin{overpic}[width=1.0\linewidth]{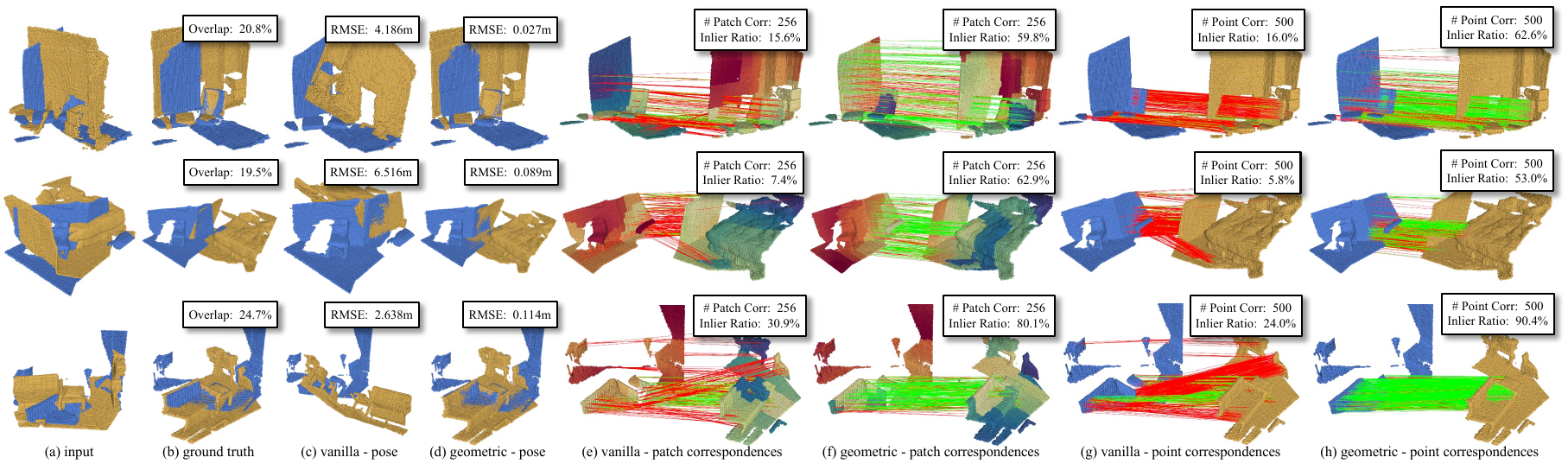}
  \end{overpic}
  \caption{\textbf{Registration results of the models with vanilla self-attention and geometric self-attention}. In the columns (e) and (f), we visualize the features of the patches with t-SNE. In the first row, the geometric self-attention helps find the inlier matches on the structure-less wall based on their geometric relationships to the more salient regions (\eg, the chairs). In the following rows, the geometric self-attention helps reject the outlier matches between the similar flat or corner patches based on their geometric relationships to the bed or the sofa.
  }
  \label{fig:ablation}
\end{figure*}

\paraspace
\ptitle{Geometric self-attention.}
To study the effectiveness of the geometric self-attention, we compare seven methods for intra-point-cloud feature learning in \cref{table:ablation-study}(a.$1$-$7$): ($1$) graph neural network~\cite{huang2021predator}, ($2$) self-attention with no positional embedding~\cite{yu2021cofinet}, ($3$) absolute coordinate embedding~\cite{sarlin2020superglue}, ($4$) relative coordinate embedding~\cite{zhao2021point}, ($5$) point pair features embedding~\cite{drost2010model,raposo2017using}, ($6$) pair-wise distance embedding, and ($7$) geometric structure embedding. Generally, injecting geometric information boosts the performance. But the gains of coordinate-based embeddings are limited due to their transformation variance. Surprisingly, GNN performs well on RR thanks to the transformation invariance of $k$NN graphs. However, it suffers from limited receptive fields which harms the IR performance. Although PPF embedding is theoretically invariant to transformation, it is hard to estimate accurate normals for the downsampled superpoints in practice, which leads to inferior performance. Our method outperforms the alternatives by a large margin on all the metrics, especially in the low-overlap scenarios, even with only the pair-wise distance embedding, demontrating the strong robustness of our method. \cref{fig:ablation} provides a gallery of the registration results of the models with vanilla self-attention and our geometric self-attention. Geometric self-attention helps infer patch matches in structure-less regions from their geometric relationships to more salient regions ($1^{\text{st}}$ row) and reject outlier matches which are similar in the feature space but different in positions ($2^{\text{nd}}$ and $3^{\text{rd}}$ rows).

\paraspace
\ptitle{Overlap-aware circle loss.}
To investigate the efficacy of the overlap-aware circle loss, we compare four loss functions for supervising the superpoint matching in \cref{table:ablation-study}(b.$1$-$4$): ($1$) cross-entropy loss~\cite{sarlin2020superglue}, ($2$) weighted cross-entropy loss~\cite{yu2021cofinet}, ($3$) circle loss~\cite{sun2020circle}, and ($4$) overlap-aware circle loss. For the first two models, an optimal transport layer is used to compute the matching matrix as in~\cite{yu2021cofinet}. Circle loss works much better than the two variants of cross-entropy loss, verifying the effectiveness of supervising superpoint matching in a metric learning fashion. Our overlap-aware circle loss beats the vanilla circle loss by a large margin on all the metrics.

\paraspace
\ptitle{Geometric structure embedding.}
Next, we study the design of geometric structure embedding. We first vary the number of nearest neighbors for computing the triplet-wise angular embedding. As shown in \cref{table:ablation-study}(c.$1$-$5$), the model with both the distance and angular embeddings outperforms the one with only the distance embedding by a large margin, which is consistent with our motivation. Moreover, increasing the number of neighbors slightly improves the performance as it provides more precise structure information, but also requires more computation. To better balance accuracy and speed, we select $k \tight{=} 3$ in our experiments unless otherwise noted.

We further investigate the influence of the temperature hyper-parameters $\sigma_d$ in \cref{eq:pde} and $\sigma_a$ in \cref{eq:tae}. From \cref{table:ablation-study}(d.$1$-$5$), the best results are achieved around the voxel size of the superpoint level (\ie, $0.2\text{m}$). A too small (where the embedding is too sensitive to distance changes) or too large (where the embedding neglects small distance variations) $\sigma_d$ could harm the performance, but the differences are not significant. And similar observations can be obtained from \cref{table:ablation-study}(e.$1$-$5$) for the angular temperature $\sigma_a$. Nevertheless, all of these models outperforms pervious methods by a large margin, indicating that GeoTransformer is still robust to the temperature hyper-parameters.

At last, we replace max pooling with average pooling when aggregating the triplet-wise angular embedding in \cref{eq:gse}. As shown in \cref{table:ablation-study}(f.$1$-$2$), max pooling performs better than average pooling. Due to self-occlusion from viewpoint changes, the nearest neighbors of a given superpoint in one point cloud could be missing in the other. Compared with average pooling, max pooling provides better robustness to the varying neighbors. For this reason, we use max pooling as the default setting.


\begin{figure}[t]
  \centering
  \begin{overpic}[width=1.0\linewidth]{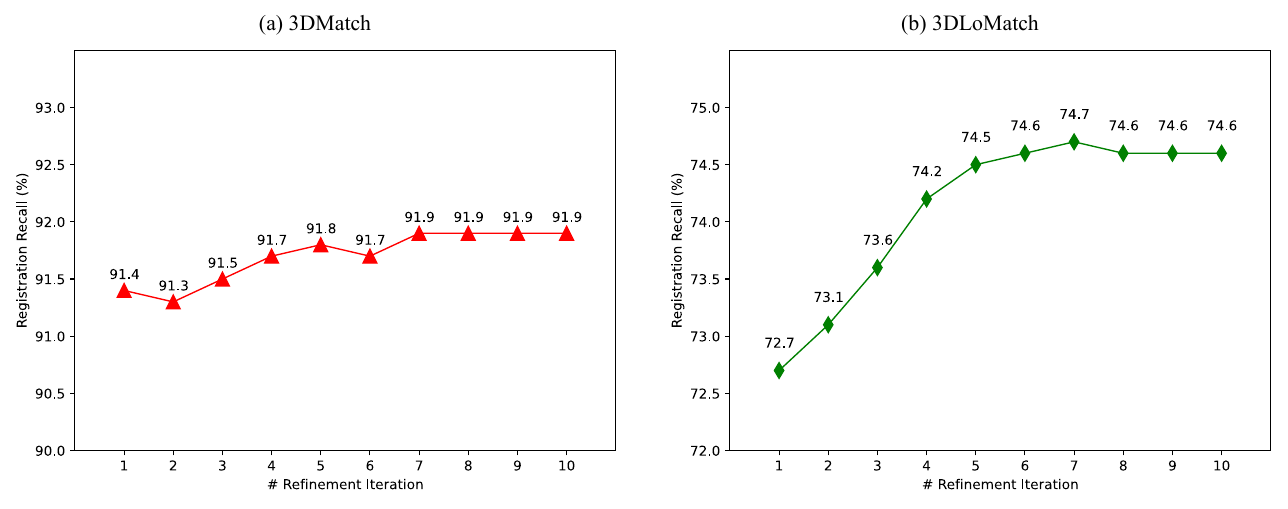}
  \end{overpic}
  \caption{\textbf{Ablation study of the pose refinement}. Pose refinement consistently improves the results and gets saturated after $5$ iterations.}
  \label{fig:refinement}
\end{figure}

\paraspace
\ptitle{Dual-normalization.}
We then investigate the effectiveness of the dual-normalization operation in the superpoint matching module. As observed in \cref{table:ablation-study}(g.$1$-$2$), it slightly improves the accuracy of the superpoint correspondences in low-overlap scenarios. As there is less overlapping context when the overlapping area is small, it is much easier to extract outlier matches between the less geometrically discriminative patches. The dual-normalization operation can mitigate this issue and slightly improves the performance.

\paraspace
\ptitle{Pose refinement.}
At last, we evaluate the impact of the pose refinement in LGR. We vary the number of refinement steps $N_r$ from $1$ to $10$. As shown in \cref{fig:refinement}, the registration recall consistently improves with more refinement iterations and quickly gets saturated. To better balance accuracy and speed, we choose $5$ iterations in the experiments.

\subsection{Comparison with Deep Robust Estimators}
\label{sec:exp-estimator}

At last, we compare GeoTransformer with recent deep robust estimators: 3DRegNet~\cite{pais20203dregnet}, DGR~\cite{choy2020deep}, PointDSC~\cite{bai2021pointdsc}, DHVR~\cite{lee2021deep} and PCAM~\cite{cao2021pcam} on 3DMatch and KITTI odometry benchmarks. For fair comparison with these methods, we follow common practice to report RTE, RRE and RR on both benchmarks. Here RR is defined as in \cref{sec:exp-outdoor} but with different thresholds. The RTE threshold is $30\text{cm}$ on 3DMatch and $60\text{cm}$ on KITTI, while the RRE threshold is $15^{\circ}$ on 3DMatch and $5^{\circ}$ on KITTI.

As shown in \cref{table:robust-estimator}, our method outperforms all the baselines by a large margin on both benchmarks. The results demonstrate the superiority of GeoTransformer over the alternative methods, although different correspondence extractors are used by those methods. It is noteworthy that our LGR is parameter-free and does not require training a specific network, which contributes to faster registration speed (0.08s of PointDSC~\cite{bai2021pointdsc} \vs 0.013s of LGR according to our experiments).


\begin{table}[!t]
\scriptsize
\centering
\caption{
\textbf{Comparison with deep robust estimators} on 3DMatch and KITTI.
\textbf{Boldfaced} numbers are the best and the second best are \underline{underlined}.
}
\label{table:robust-estimator}
\begin{tabular}{l|ccc}
\toprule
Model & RTE(cm) & RRE($^{\circ}$) & RR(\%) \\
\midrule
\multicolumn{4}{c}{3DMatch} \\
\midrule
FCGF+3DRegNet~\cite{pais20203dregnet} & 8.13 & 2.74 & 77.8 \\
FCGF+DGR~\cite{choy2020deep} & 7.36 & 2.33 & 86.5 \\
FCGF+PointDSC~\cite{bai2021pointdsc} & \underline{6.55} & \underline{2.06} & \underline{93.3} \\
FCGF+DHVR~\cite{lee2021deep} & 6.61 & 2.08 & 91.4 \\
PCAM~\cite{cao2021pcam} & $\sim$7 & 2.16 & 92.4 \\
GeoTransformer (\emph{ours}, LGR) & \textbf{5.69} & \textbf{1.92} & \textbf{95.7} \\
\midrule
\multicolumn{4}{c}{3DLoMatch} \\
\midrule
FCGF+PointDSC~\cite{bai2021pointdsc} & \underline{10.50} & \underline{3.82} & \underline{56.2} \\
FCGF+DHVR~\cite{lee2021deep} & 11.76 & 3.88 & 55.6 \\
GeoTransformer (\emph{ours}, LGR) & \textbf{8.55} & \textbf{2.95} & \textbf{78.0} \\
\midrule
\multicolumn{4}{c}{KITTI} \\
\midrule
FCGF+DGR~\cite{choy2020deep} & 21.7 & 0.34 & 96.9 \\
FCGF+PointDSC~\cite{bai2021pointdsc} & 20.9 & 0.33 & 98.2 \\
FCGF+DHVR~\cite{lee2021deep} & 19.8 & \underline{0.29} & \underline{99.1} \\
PCAM~\cite{cao2021pcam} & \underline{$\sim$8} & 0.33 & 97.2 \\
GeoTransformer (\emph{ours}, LGR) & \textbf{6.5} & \textbf{0.24} & \textbf{99.5} \\
\bottomrule
\end{tabular}
\end{table}


\section{Conclusion}

We have presented Geometric Transformer to learn robust coarse-to-fine correspondences for point cloud registration. Through encoding pair-wise distances and triplet-wise angles among superpoints, our method captures the geometric consistency across point clouds with transformation invariance.
Thanks to the reliable correspondences, it attains fast and accurate registration in a RANSAC-free manner.
Extensive experiments on five challenging benchmarks have demonstrated the efficacy of GeoTransformer.

\paraspace
\ptitle{Limitations.}
In spite of the state-of-the-art performance, there are still some limitations in GeoTransformer.
(1) GeoTransformer relies on uniformly downsampled superpoints to hierarchically extract correspondences. However, there could be numerous superpoints if the input point clouds cover a large area, which could cause huge memory footprint and computational cost. For this reason, we might need to carefully select the downsampling rate to balance performance and efficiency. For example, we add an additional downsampling stage on KITTI and Augmented ICL-NUIM, which effectively improves the memory and computational efficiency without sacrificing the accuracy.
(2) The inflexibility of uniformly sampling superpoints (patches) is another concern. In practice, it is common that a single object is decomposed into multiple patches, but it could be easily registered as a whole. So we believe that it is a very promising topic to integrate point cloud registration with semantic scene understanding tasks (\eg, object detection and instance segmentation), which converts scene registration into semantic object registration.

\paraspace
\ptitle{Future work.}
Besides the aforementioned limitations, there are also many directions where GeoTransformer could be extended, including cross-modality (\eg, 2D-3D) registration and end-to-end non-rigid registration.


\ifCLASSOPTIONcompsoc
  \section*{Acknowledgments}
\else
  \section*{Acknowledgment}
\fi

This work was supported in part by the NSFC (62132021, 62102435) and the National Key Research and Development Program of China (2018AAA0102200).

\ifCLASSOPTIONcaptionsoff
  \newpage
\fi



%

\bibliographystyle{IEEEtran}
\bibliography{geotrans}

\begin{thebibliography}{10}
\providecommand{\url}[1]{#1}
\csname url@samestyle\endcsname
\providecommand{\newblock}{\relax}
\providecommand{\bibinfo}[2]{#2}
\providecommand{\BIBentrySTDinterwordspacing}{\spaceskip=0pt\relax}
\providecommand{\BIBentryALTinterwordstretchfactor}{4}
\providecommand{\BIBentryALTinterwordspacing}{\spaceskip=\fontdimen2\font plus
\BIBentryALTinterwordstretchfactor\fontdimen3\font minus
  \fontdimen4\font\relax}
\providecommand{\BIBforeignlanguage}[2]{{%
\expandafter\ifx\csname l@#1\endcsname\relax
\typeout{** WARNING: IEEEtran.bst: No hyphenation pattern has been}%
\typeout{** loaded for the language `#1'. Using the pattern for}%
\typeout{** the default language instead.}%
\else
\language=\csname l@#1\endcsname
\fi
#2}}
\providecommand{\BIBdecl}{\relax}
\BIBdecl

\bibitem{deng2018ppfnet}
H.~Deng, T.~Birdal, and S.~Ilic, ``Ppfnet: Global context aware local features
  for robust 3d point matching,'' in \emph{CVPR}, 2018, pp. 195--205.

\bibitem{gojcic2019perfect}
Z.~Gojcic, C.~Zhou, J.~D. Wegner, and A.~Wieser, ``The perfect match: 3d point
  cloud matching with smoothed densities,'' in \emph{CVPR}, 2019, pp.
  5545--5554.

\bibitem{choy2019fully}
C.~Choy, J.~Park, and V.~Koltun, ``Fully convolutional geometric features,'' in
  \emph{CVPR}, 2019, pp. 8958--8966.

\bibitem{bai2020d3feat}
X.~Bai, Z.~Luo, L.~Zhou, H.~Fu, L.~Quan, and C.-L. Tai, ``D3feat: Joint
  learning of dense detection and description of 3d local features,'' in
  \emph{CVPR}, 2020, pp. 6359--6367.

\bibitem{huang2021predator}
S.~Huang, Z.~Gojcic, M.~Usvyatsov, A.~Wieser, and K.~Schindler, ``Predator:
  Registration of 3d point clouds with low overlap,'' in \emph{CVPR}, 2021, pp.
  4267--4276.

\bibitem{yu2021cofinet}
H.~Yu, F.~Li, M.~Saleh, B.~Busam, and S.~Ilic, ``Cofinet: Reliable
  coarse-to-fine correspondences for robust pointcloud registration,''
  \emph{NeurIPS}, vol.~34, 2021.

\bibitem{ao2021spinnet}
S.~Ao, Q.~Hu, B.~Yang, A.~Markham, and Y.~Guo, ``Spinnet: Learning a general
  surface descriptor for 3d point cloud registration,'' in \emph{CVPR}, 2021,
  pp. 11\,753--11\,762.

\bibitem{rocco2018neighbourhood}
I.~Rocco, M.~Cimpoi, R.~Arandjelovi{\'c}, A.~Torii, T.~Pajdla, and J.~Sivic,
  ``Neighbourhood consensus networks,'' \emph{NeurIPS}, vol.~31, pp.
  1651--1662, 2018.

\bibitem{zhou2021patch2pix}
Q.~Zhou, T.~Sattler, and L.~Leal-Taixe, ``Patch2pix: Epipolar-guided
  pixel-level correspondences,'' in \emph{CVPR}, 2021, pp. 4669--4678.

\bibitem{sun2021loftr}
J.~Sun, Z.~Shen, Y.~Wang, H.~Bao, and X.~Zhou, ``Loftr: Detector-free local
  feature matching with transformers,'' in \emph{CVPR}, 2021, pp. 8922--8931.

\bibitem{vaswani2017attention}
A.~Vaswani, N.~Shazeer, N.~Parmar, J.~Uszkoreit, L.~Jones, A.~N. Gomez,
  {\L}.~Kaiser, and I.~Polosukhin, ``Attention is all you need,'' in
  \emph{NeurIPS}, 2017, pp. 5998--6008.

\bibitem{wang2019deep}
Y.~Wang and J.~M. Solomon, ``Deep closest point: Learning representations for
  point cloud registration,'' in \emph{ICCV}, 2019, pp. 3523--3532.

\bibitem{zhao2021point}
H.~Zhao, L.~Jiang, J.~Jia, P.~H. Torr, and V.~Koltun, ``Point transformer,'' in
  \emph{ICCV}, 2021, pp. 16\,259--16\,268.

\bibitem{yang2019modeling}
J.~Yang, Q.~Zhang, B.~Ni, L.~Li, J.~Liu, M.~Zhou, and Q.~Tian, ``Modeling point
  clouds with self-attention and gumbel subset sampling,'' in \emph{CVPR},
  2019, pp. 3323--3332.

\bibitem{sarlin2020superglue}
P.-E. Sarlin, D.~DeTone, T.~Malisiewicz, and A.~Rabinovich, ``Superglue:
  Learning feature matching with graph neural networks,'' in \emph{CVPR}, 2020,
  pp. 4938--4947.

\bibitem{zeng20173dmatch}
A.~Zeng, S.~Song, M.~Nie{\ss}ner, M.~Fisher, J.~Xiao, and T.~Funkhouser,
  ``3dmatch: Learning local geometric descriptors from rgb-d reconstructions,''
  in \emph{CVPR}, 2017, pp. 1802--1811.

\bibitem{geiger2012we}
A.~Geiger, P.~Lenz, and R.~Urtasun, ``Are we ready for autonomous driving? the
  kitti vision benchmark suite,'' in \emph{CVPR}.\hskip 1em plus 0.5em minus
  0.4em\relax IEEE, 2012, pp. 3354--3361.

\bibitem{choi2015robust}
S.~Choi, Q.-Y. Zhou, and V.~Koltun, ``Robust reconstruction of indoor scenes,''
  in \emph{CVPR}, 2015, pp. 5556--5565.

\bibitem{li2022lepard}
Y.~Li and T.~Harada, ``Lepard: Learning partial point cloud matching in rigid
  and deformable scenes,'' in \emph{CVPR}, 2022, pp. 5554--5564.

\bibitem{qin2022geometric}
Z.~Qin, H.~Yu, C.~Wang, Y.~Guo, Y.~Peng, and K.~Xu, ``Geometric transformer for
  fast and robust point cloud registration,'' in \emph{CVPR (oral)}, 2022, pp.
  11\,143--11\,152.

\bibitem{deng2018ppf}
H.~Deng, T.~Birdal, and S.~Ilic, ``Ppf-foldnet: Unsupervised learning of
  rotation invariant 3d local descriptors,'' in \emph{ECCV}, 2018, pp.
  602--618.

\bibitem{wang2022you}
H.~Wang, Y.~Liu, Z.~Dong, W.~Wang, and B.~Yang, ``You only hypothesize once:
  Point cloud registration with rotation-equivariant descriptors,'' in
  \emph{ACM MM}, 2022, pp. 1630--1641.

\bibitem{wang2019prnet}
Y.~Wang and J.~Solomon, ``Prnet: self-supervised learning for
  partial-to-partial registration,'' in \emph{NeurIPS}, 2019, pp. 8814--8826.

\bibitem{yuan2020deepgmr}
W.~Yuan, B.~Eckart, K.~Kim, V.~Jampani, D.~Fox, and J.~Kautz, ``Deepgmr:
  Learning latent gaussian mixture models for registration,'' in
  \emph{ECCV}.\hskip 1em plus 0.5em minus 0.4em\relax Springer, 2020, pp.
  733--750.

\bibitem{li2020iterative}
J.~Li, C.~Zhang, Z.~Xu, H.~Zhou, and C.~Zhang, ``Iterative distance-aware
  similarity matrix convolution with mutual-supervised point elimination for
  efficient point cloud registration,'' in \emph{ECCV}.\hskip 1em plus 0.5em
  minus 0.4em\relax Springer, 2020, pp. 378--394.

\bibitem{yew2020rpm}
Z.~J. Yew and G.~H. Lee, ``Rpm-net: Robust point matching using learned
  features,'' in \emph{CVPR}, 2020, pp. 11\,824--11\,833.

\bibitem{fu2021robust}
K.~Fu, S.~Liu, X.~Luo, and M.~Wang, ``Robust point cloud registration framework
  based on deep graph matching,'' in \emph{CVPR}, 2021, pp. 8893--8902.

\bibitem{zhang2022end}
Z.~Zhang, J.~Sun, Y.~Dai, D.~Zhou, X.~Song, and M.~He, ``End-to-end learning
  the partial permutation matrix for robust 3d point cloud registration,'' in
  \emph{AAAI}, vol.~36, no.~3, 2022, pp. 3399--3407.

\bibitem{besl1992method}
P.~J. Besl and N.~D. McKay, ``Method for registration of 3-d shapes,'' in
  \emph{Sensor fusion IV: control paradigms and data structures}, vol.
  1611.\hskip 1em plus 0.5em minus 0.4em\relax International Society for Optics
  and Photonics, 1992, pp. 586--606.

\bibitem{aoki2019pointnetlk}
Y.~Aoki, H.~Goforth, R.~A. Srivatsan, and S.~Lucey, ``Pointnetlk: Robust \&
  efficient point cloud registration using pointnet,'' in \emph{CVPR}, 2019,
  pp. 7163--7172.

\bibitem{huang2020feature}
X.~Huang, G.~Mei, and J.~Zhang, ``Feature-metric registration: A fast
  semi-supervised approach for robust point cloud registration without
  correspondences,'' in \emph{CVPR}, 2020, pp. 11\,366--11\,374.

\bibitem{xu2021omnet}
H.~Xu, S.~Liu, G.~Wang, G.~Liu, and B.~Zeng, ``Omnet: Learning overlapping mask
  for partial-to-partial point cloud registration,'' in \emph{ICCV}, 2021, pp.
  3132--3141.

\bibitem{xu2022finet}
H.~Xu, N.~Ye, G.~Liu, B.~Zeng, and S.~Liu, ``Finet: Dual branches feature
  interaction for partial-to-partial point cloud registration,'' in
  \emph{AAAI}, vol.~36, no.~3, 2022, pp. 2848--2856.

\bibitem{pais20203dregnet}
G.~D. Pais, S.~Ramalingam, V.~M. Govindu, J.~C. Nascimento, R.~Chellappa, and
  P.~Miraldo, ``3dregnet: A deep neural network for 3d point registration,'' in
  \emph{CVPR}, 2020, pp. 7193--7203.

\bibitem{choy2020deep}
C.~Choy, W.~Dong, and V.~Koltun, ``Deep global registration,'' in \emph{CVPR},
  2020, pp. 2514--2523.

\bibitem{bai2021pointdsc}
X.~Bai, Z.~Luo, L.~Zhou, H.~Chen, L.~Li, Z.~Hu, H.~Fu, and C.-L. Tai,
  ``Pointdsc: Robust point cloud registration using deep spatial consistency,''
  in \emph{CVPR}, 2021, pp. 15\,859--15\,869.

\bibitem{rusu2009fast}
R.~B. Rusu, N.~Blodow, and M.~Beetz, ``Fast point feature histograms (fpfh) for
  3d registration,'' in \emph{ICRA}.\hskip 1em plus 0.5em minus 0.4em\relax
  IEEE, 2009, pp. 3212--3217.

\bibitem{drost2010model}
B.~Drost, M.~Ulrich, N.~Navab, and S.~Ilic, ``Model globally, match locally:
  Efficient and robust 3d object recognition,'' in \emph{CVPR}.\hskip 1em plus
  0.5em minus 0.4em\relax IEEE, 2010, pp. 998--1005.

\bibitem{raposo2017using}
C.~Raposo and J.~P. Barreto, ``Using 2 point+ normal sets for fast registration
  of point clouds with small overlap,'' in \emph{ICRA}.\hskip 1em plus 0.5em
  minus 0.4em\relax IEEE, 2017, pp. 5652--5658.

\bibitem{leordeanu2005spectral}
M.~Leordeanu and M.~Hebert, ``A spectral technique for correspondence problems
  using pairwise constraints,'' in \emph{ICCV}, vol.~2.\hskip 1em plus 0.5em
  minus 0.4em\relax IEEE, 2005, pp. 1482--1489.

\bibitem{yang2020teaser}
H.~Yang, J.~Shi, and L.~Carlone, ``Teaser: Fast and certifiable point cloud
  registration,'' \emph{IEEE Transactions on Robotics}, vol.~37, no.~2, pp.
  314--333, 2020.

\bibitem{thomas2019kpconv}
H.~Thomas, C.~R. Qi, J.-E. Deschaud, B.~Marcotegui, F.~Goulette, and L.~J.
  Guibas, ``Kpconv: Flexible and deformable convolution for point clouds,'' in
  \emph{ICCV}, 2019, pp. 6411--6420.

\bibitem{lin2017feature}
T.-Y. Lin, P.~Doll{\'a}r, R.~Girshick, K.~He, B.~Hariharan, and S.~Belongie,
  ``Feature pyramid networks for object detection,'' in \emph{CVPR}, 2017, pp.
  2117--2125.

\bibitem{li2018so}
J.~Li, B.~M. Chen, and G.~H. Lee, ``So-net: Self-organizing network for point
  cloud analysis,'' in \emph{CVPR}, 2018, pp. 9397--9406.

\bibitem{dosovitskiy2020image}
A.~Dosovitskiy, L.~Beyer, A.~Kolesnikov, D.~Weissenborn, X.~Zhai,
  T.~Unterthiner, M.~Dehghani, M.~Minderer, G.~Heigold, S.~Gelly \emph{et~al.},
  ``An image is worth 16x16 words: Transformers for image recognition at
  scale,'' 2021.

\bibitem{sinkhorn1967concerning}
R.~Sinkhorn and P.~Knopp, ``Concerning nonnegative matrices and doubly
  stochastic matrices,'' \emph{Pacific Journal of Mathematics}, vol.~21, no.~2,
  pp. 343--348, 1967.

\bibitem{sun2020circle}
Y.~Sun, C.~Cheng, Y.~Zhang, C.~Zhang, L.~Zheng, Z.~Wang, and Y.~Wei, ``Circle
  loss: A unified perspective of pair similarity optimization,'' in
  \emph{CVPR}, 2020, pp. 6398--6407.

\bibitem{wu20153d}
Z.~Wu, S.~Song, A.~Khosla, F.~Yu, L.~Zhang, X.~Tang, and J.~Xiao, ``3d
  shapenets: A deep representation for volumetric shapes,'' in \emph{CVPR},
  2015, pp. 1912--1920.

\bibitem{paszke2019pytorch}
A.~Paszke, S.~Gross, F.~Massa, A.~Lerer, J.~Bradbury, G.~Chanan, T.~Killeen,
  Z.~Lin, N.~Gimelshein, L.~Antiga \emph{et~al.}, ``Pytorch: An imperative
  style, high-performance deep learning library,'' \emph{NeurIPS}, vol.~32, pp.
  8026--8037, 2019.

\bibitem{kingma2014adam}
D.~P. Kingma and J.~Ba, ``Adam: A method for stochastic optimization,'' 2015.

\bibitem{yew20183dfeat}
Z.~J. Yew and G.~H. Lee, ``3dfeat-net: Weakly supervised local 3d features for
  point cloud registration,'' in \emph{ECCV}, 2018, pp. 607--623.

\bibitem{lu2021hregnet}
F.~Lu, G.~Chen, Y.~Liu, L.~Zhang, S.~Qu, S.~Liu, and R.~Gu, ``Hregnet: A
  hierarchical network for large-scale outdoor lidar point cloud
  registration,'' in \emph{ICCV}, 2021, pp. 16\,014--16\,023.

\bibitem{qi2017pointnet}
C.~R. Qi, H.~Su, K.~Mo, and L.~J. Guibas, ``Pointnet: Deep learning on point
  sets for 3d classification and segmentation,'' in \emph{Proceedings of the
  IEEE conference on computer vision and pattern recognition}, 2017, pp.
  652--660.

\bibitem{zhou2016fast}
Q.-Y. Zhou, J.~Park, and V.~Koltun, ``Fast global registration,'' in
  \emph{ECCV}.\hskip 1em plus 0.5em minus 0.4em\relax Springer, 2016, pp.
  766--782.

\bibitem{fischler1981random}
M.~A. Fischler and R.~C. Bolles, ``Random sample consensus: a paradigm for
  model fitting with applications to image analysis and automated
  cartography,'' \emph{Communications of the ACM}, vol.~24, no.~6, pp.
  381--395, 1981.

\bibitem{handa2014benchmark}
A.~Handa, T.~Whelan, J.~McDonald, and A.~J. Davison, ``A benchmark for rgb-d
  visual odometry, 3d reconstruction and slam,'' in \emph{ICRA}.\hskip 1em plus
  0.5em minus 0.4em\relax IEEE, 2014, pp. 1524--1531.

\bibitem{kummerle2011g}
R.~K{\"u}mmerle, G.~Grisetti, H.~Strasdat, K.~Konolige, and W.~Burgard,
  ``g$^2$o: A general framework for graph optimization,'' in \emph{ICRA}.\hskip
  1em plus 0.5em minus 0.4em\relax IEEE, 2011, pp. 3607--3613.

\bibitem{zhou2018open3d}
Q.-Y. Zhou, J.~Park, and V.~Koltun, ``Open3d: A modern library for 3d data
  processing,'' \emph{arXiv preprint arXiv:1801.09847}, 2018.

\bibitem{li20214dcomplete}
Y.~Li, H.~Takehara, T.~Taketomi, B.~Zheng, and M.~Nie{\ss}ner, ``4dcomplete:
  Non-rigid motion estimation beyond the observable surface,'' in \emph{ICCV},
  2021, pp. 12\,706--12\,716.

\bibitem{li2008global}
H.~Li, R.~W. Sumner, and M.~Pauly, ``Global correspondence optimization for
  non-rigid registration of depth scans,'' in \emph{Computer graphics forum},
  vol.~27, no.~5.\hskip 1em plus 0.5em minus 0.4em\relax Wiley Online Library,
  2008, pp. 1421--1430.

\bibitem{yao2020quasi}
Y.~Yao, B.~Deng, W.~Xu, and J.~Zhang, ``Quasi-newton solver for robust
  non-rigid registration,'' in \emph{Proceedings of the IEEE/CVF conference on
  computer vision and pattern recognition}, 2020, pp. 7600--7609.

\bibitem{lee2021deep}
J.~Lee, S.~Kim, M.~Cho, and J.~Park, ``Deep hough voting for robust global
  registration,'' in \emph{ICCV}, 2021, pp. 15\,994--16\,003.

\bibitem{cao2021pcam}
A.-Q. Cao, G.~Puy, A.~Boulch, and R.~Marlet, ``Pcam: Product of cross-attention
  matrices for rigid registration of point clouds,'' in \emph{ICCV}, 2021, pp.
  13\,229--13\,238.

\end{thebibliography}


%

\begin{IEEEbiography}[{\includegraphics[width=1in,height=1.25in,clip,keepaspectratio]{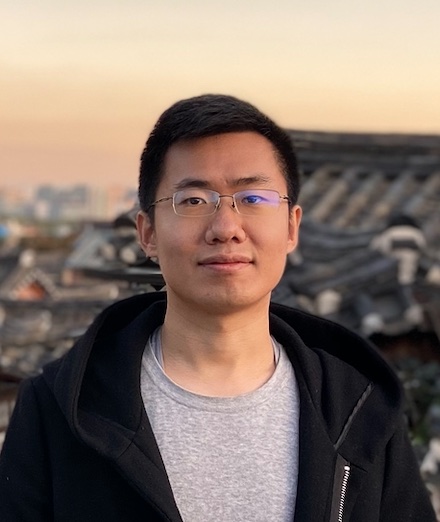}}]{Zheng Qin}
received the B.E. and M.E. degree in computer science and technology from National University of Defense Technology (NUDT), China, in 2016 and 2018, respectively, where he is currently pursuing the Ph.D. degree. His research interests focus on 3D vision, including point cloud registration, pose estimation, and 3D representation learning.
\end{IEEEbiography}

\begin{IEEEbiography}[{\includegraphics[width=1in,height=1.25in,clip,keepaspectratio]{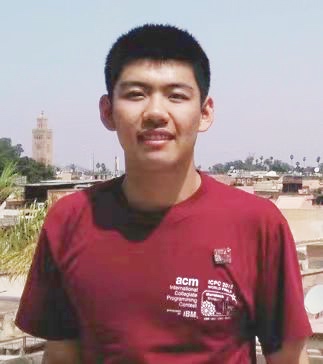}}]{Hao Yu}
is a PhD student at the Chair for Computer Aided Medical Procedures \& Augmented Reality (CAMP) of TU Munich under supervision of PD. Dr. Slobodan Ilic and Dr. Benjamin Busam. He recieved his Master's degree in Computer Science from National University of Defense Technology, China, where he also completed his Bachelor's study in Network Engineering. His research interest includes 3D local descriptors and point cloud registration.
\end{IEEEbiography}

\begin{IEEEbiography}[{\includegraphics[width=1in,height=1.25in,clip,keepaspectratio]{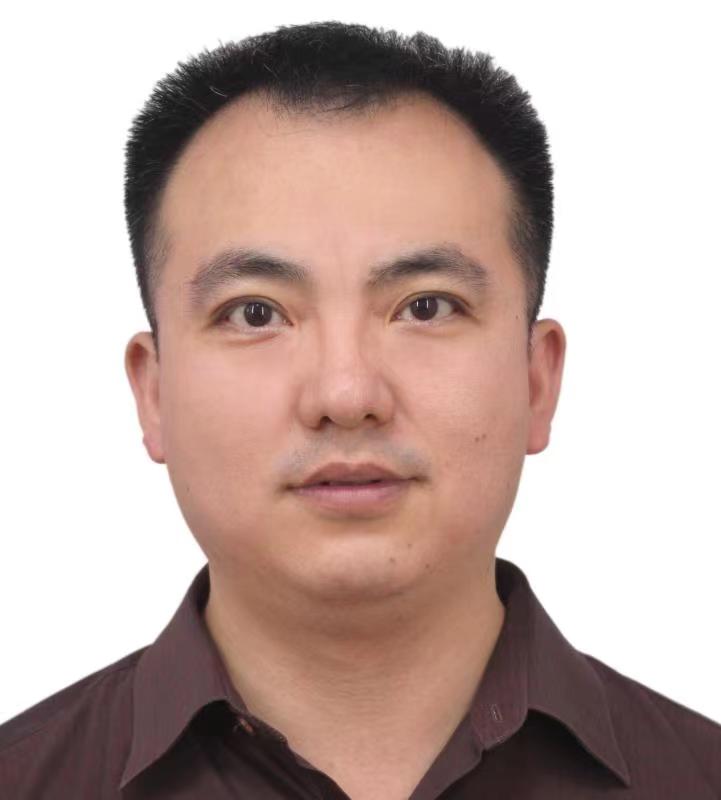}}]{Changjian Wang}
received his Ph.D. degree in computer science from the School of Computer, National University of Defense Technology. He is currently an Associate Professor of the National University of Defense Technology (NUDT), Changsha, China. His current research interests include medical image analysis, natural language processing and big data.
\end{IEEEbiography}

\begin{IEEEbiography}[{\includegraphics[width=1in,height=1.25in,clip,keepaspectratio]{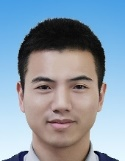}}]{Yulan Guo}
received the B.E. and Ph.D. degrees from National University of Defense Technology (NUDT) in 2008 and 2015, respectively. He has authored over 100 articles at highly referred journals and conferences. His current research interests focus on 3D vision, particularly on 3D feature learning, 3D modeling, 3D object recognition, and scene understanding. He served as an associate editor for IEEE Transactions on Image Processing, IET Computer Vision, IET Image Processing, and Computers \& Graphics. He also served as an area chair for CVPR 2021, ICCV 2021, and ACM Multimedia 2021. He organized several tutorials, workshops, and challenges in prestigious conferences, such as CVPR 2016, CVPR 2019, ICCV 2021, 3DV 2021, CVPR 2022, ICPR 2022, and ECCV 2022. He is a Senior Member of IEEE and ACM.
\end{IEEEbiography}

\begin{IEEEbiography}[{\includegraphics[width=1in,height=1.25in,clip,keepaspectratio]{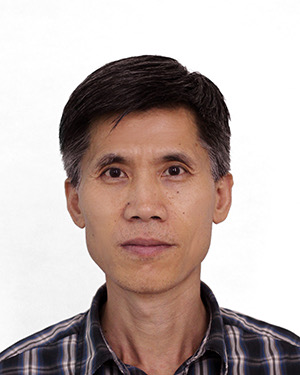}}]{Yuxing Peng}
received the Ph.D. degree from the National University of Defense Technology (NUDT), China, in 1996. He is currently a Professor with the College of Computer Science and Technology, NUDT. His research interests primarily focus on machine learning, data analysis and cloud computing.
\end{IEEEbiography}

\begin{IEEEbiography}[{\includegraphics[width=1in,height=1.25in,clip,keepaspectratio]{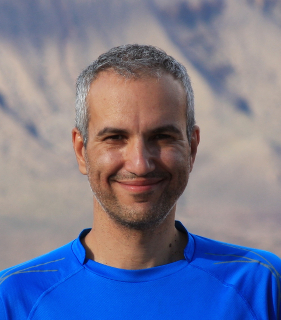}}]{Slobodan Ilic}
is currently senior key expert research scientist at Siemens Corporate Technology in Munich, Perlach. He is also a visiting researcher and lecturer at Computer Science Department of TUM and closely works with the CAMP Chair. From 2009 until end of 2013 he was leading the Computer Vision Group of CAMP at TUM, and before that he was a senior researcher at Deutsche Telekom Laboratories in Berlin. In 2005 he obtained his PhD at EPFL in Switzerland under supervision of Pascal Fua. His research interests include: 3D reconstruction, deformable surface modelling and tracking, real-time object detection and tracking, human pose estimation and semantic segmentation.
\end{IEEEbiography}

\begin{IEEEbiography}[{\includegraphics[width=1in,height=1.25in,clip,keepaspectratio]{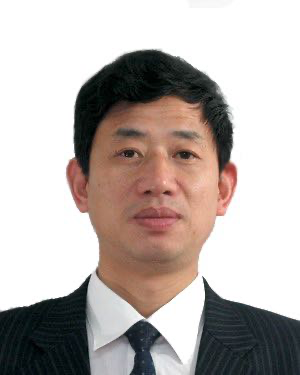}}]{Dewen Hu}
received the B.Sc. and M.Sc. degrees from Xi’an Jiaotong University, China, in 1983 and 1986, respectively, and the Ph.D. degree from the National University of Defense Technology, in 1999. In 1986, he was with the National University of Defense Technology. From October 1995 to October 1996, he was a Visiting Scholar with The University of Sheffield, U.K. In 1996, he was promoted as a Professor. He has authored more than 200 articles in journals, such as the Brain, the Proceedings of the National Academy of Sciences of the United States of America, the NeuroImage, the Human Brain Mapping, the IEEE Transactions on Pattern Analysis and Machine Intelligence, the IEEE Transactions on Image Processing, the IEEE Transactions on Signal Processing, the IEEE Transactions on Neural Networks and Learning Systems, the IEEE Transactions on Medical Imaging, and the IEEE Transactions on Biomedical Engineering. His research interests include pattern recognition and cognitive neuroscience. He is currently an Action Editor of Neural Networks, and an Associate Editor of IEEE Transactions on Systems, Man, and Cybernetics: Systems.
\end{IEEEbiography}

\begin{IEEEbiography}[{\includegraphics[width=1in,height=1.25in,clip,keepaspectratio]{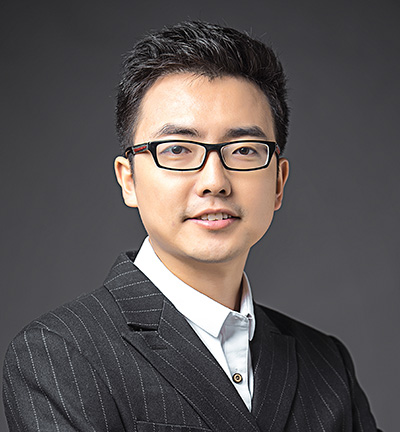}}]{Kai Xu}
is a Professor at the College of Computer, NUDT, where he received his Ph.D. in 2011. He conducted visiting research at Simon Fraser University and Princeton University. His research interests include geometric modeling and shape analysis, especially on data-driven approaches to the problems in those directions, as well as 3D vision and its robotic applications. He has published over 80 research papers, including 20+ SIGGRAPH/TOG papers. He has co-organized several SIGGRAPH Asia courses and Eurographics STAR tutorials. He serves on the editorial board of ACM Transactions on Graphics, Computer Graphics Forum, Computers \& Graphics, and The Visual Computer. He also served as program co-chair of CAD/Graphics 2017, ICVRV 2017 and ISVC 2018, as well as PC member for several prestigious conferences including SIGGRAPH, SIGGRAPH Asia, Eurographics, SGP, PG, etc. His research work can be found in his personal website: www.kevinkaixu.net.
\end{IEEEbiography}


\vfill


\end{document}